\documentclass[letterpaper]{article} 
\usepackage{aaai24}  
\usepackage{times}  
\usepackage{helvet}  
\usepackage{courier}  
\usepackage[hyphens]{url}  
\usepackage{graphicx} 
\usepackage{natbib}  
\usepackage{caption} 
\usepackage{algorithm}
\usepackage{algorithmicx}
\usepackage[noend]{algpseudocode}
\usepackage{newfloat}
\usepackage{listings}
\DeclareCaptionStyle{ruled}{labelfont=normalfont,labelsep=colon,strut=off} 
\floatstyle{ruled}
\newfloat{listing}{tb}{lst}{}
\floatname{listing}{Listing}
\nocopyright 

\usepackage{nicematrix}
\usepackage{dsfont}
\usepackage{amssymb}
\usepackage{multirow}
\usepackage{tikz}
\usepackage{acronym}
\newacro{FFT}{fast Fourier transform}
\newacro{CSPs}{Constraint-satisfaction problems}
\newacro{SAT}{Boolean satisfiability}
\newacro{CNF}{conjunctive normal form}
\newacro{CLS}{continuous local search}
\newacro{SOTA}{state-of-the-art}
\newacro{CDCL}{conflict-driven clause learning}
\newacro{DLS}{Discrete local search}
\newacro{ESPs}{elementary symmetric polynomials}
\newacro{CTDS}{continuous-time deterministic solver}
\newacro{BDD}{binary decision diagram}
\newacro{DFT}{discrete Fourier transform}
\newacro{ERWA}{exponential recency weighted average}
\newacro{VBS}{virtual best solver}
\newacro{MSE}{MaxSAT Evaluations}
\newacro{ML}{machine learning}
\newacro{DFT}{discrete Fourier transform}
\newacro{PGFP}{proximal gradient fixed-point}
\newacro{LHS}{left hand side}
\newacro{RHS}{right hand side}

\newtheorem{theorem}{Theorem}
\newtheorem{lemma}{Lemma}
\newtheorem{corollary}{Corollary}
\newtheorem{definition}{Definition}
\newtheorem{proposition}{Proposition}
\newtheorem{example}{Example}
\newtheorem{fact}{Fact}

\title{Massively Parallel Continuous Local Search for Hybrid SAT Solving on GPUs}
\author {
    Yunuo Cen,\textsuperscript{\rm 1}
    Zhiwei Zhang, \textsuperscript{\rm 2}
    Xuanyao Fong \textsuperscript{\rm 1}
}
\affiliations {
    \textsuperscript{\rm 1} Department of Electrical and Computer Engineering, National University of Singapore\\
    \textsuperscript{\rm 2} Department of Computer Science, Rice University\\
    cenyunuo@u.nus.edu, zhiwei@rice.edu, kelvin.xy.fong@nus.edu.sg
}

\begin{document}

\maketitle

\begin{abstract}
Although \ac{SOTA} SAT solvers based on conflict-driven clause learning (CDCL) have achieved remarkable engineering success, their sequential nature limits the parallelism that may be extracted for acceleration on platforms such as the graphics processing unit (GPU).
In this work, we propose \texttt{FastFourierSAT}, a highly parallel hybrid SAT solver based on gradient-driven continuous local search (CLS). 
This is realized by a novel parallel algorithm inspired by the Fast Fourier Transform (FFT)-based convolution for computing the elementary symmetric polynomials (ESPs), which is the major computational task in previous CLS methods. 
The complexity of our algorithm matches the best previous result. 
Furthermore, the substantial parallelism inherent in our algorithm can leverage the GPU for acceleration, demonstrating significant improvement over the previous CLS approaches.
We also propose to incorporate the restart heuristics in CLS to improve search efficiency. 
We compare our approach with the \ac{SOTA} parallel SAT solvers on several benchmarks. 
Our results show that \texttt{FastFourierSAT} computes the gradient 100+ times faster than previous prototypes implemented on CPU.
Moreover, \texttt{FastFourierSAT} solves most instances and demonstrates promising performance on larger-size instances.

\end{abstract}

\section{Introduction}

\ac{CSPs} are fundamental in mathematics, physics, and computer science. 
The \ac{SAT} problem is a paradigmatic class of \ac{CSPs}, where each variable takes values from the binary set \{\texttt{True, False}\}. 
Solving SAT efficiently is of utmost significance in computer science, both from a theoretical and a practical perspective~\cite{kyrillidis2020fouriersat}. 
The dominating technique of SAT has evolved from local search~\cite{mitchell1992new} to DPLL~\cite{davis1960computing, davis1962machine} and \ac{CDCL}~\cite{marques1999grasp}. 
\ac{CDCL} SAT solvers have been highly successful, constantly solving industrial benchmarks with millions of variables. 
Numerous problems in various domains are encoded and tackled by SAT solving, e.g., information theory~\cite{golia2022scalable}, VLSI design~\cite{wang2023fastpass}, and quantum computing~\cite{vardi2023solving}. 

Despite the domination of \ac{CDCL} SAT solvers, most of them rely on a sequential process, where each decision and propagation step depends on the history~\cite{hamadi2013seven}. 
The sequential nature of \ac{CDCL} solvers makes it challenging to natively parallelize them and leverage advanced computational resources such as multicore CPU, GPU, and TPU~\cite{jouppi2017datacenter}.
Most parallel SAT solvers are based on divide-and-conquer and portfolio principles, which only exploit \emph{thread-level} parallelism~\cite{martins2012overview}. 
A guiding path is usually used to divide the search space into disjoint subspaces, which are individually searched in parallel for a solution using multiple threads~\cite{zhang1996psato}. 
Different single-thread solvers use different search configurations to diversify the search trajectories~\cite{hamadi2010manysat}. 
In modern SAT solvers, threads may communicate with each other to exchange information~\cite{le2017painless}. 
However, the sequential dependency limits the opportunities for instruction-level parallelism.

Some SAT solvers achieve \emph{instruction-level} parallelism to some extent by incorporating data structures that allow for efficient clause database management, or conflict analysis. 
\texttt{CUD@SAT}~\cite{dal2015cud} proposed a parallel unit propagation algorithm on GPU but gain limited acceleration. 
\texttt{GPUShareSat}~\cite{prevot2021leveraging} uses GPU for parallel clause checking, thereby minimizing the shared database of a portfolio approach. 
\texttt{ParaFROST}~\cite{osama2021sat} uses GPU to implement the parallel inprocessing to simplify the database.

Motivated by the success in efficiently training neural networks, it is promising to exploit the potential of hardware advancement in \ac{ML}, such as GPU, in SAT solving at thread, instruction, and \emph{data-level}. 
In this paper, we propose a highly parallelized SAT solver based on \ac{CLS}.  
Initiated by \texttt{FourierSAT}~\cite{kyrillidis2020fouriersat} and followed by \texttt{GradSAT}~\cite{kyrillidis2021continuous}, \ac{CLS}-based SAT solvers have been recently actively studied as a novel line of SAT framework. 
These solvers transform SAT into a polynomial optimization problem over the real domain and apply gradient-based optimizers. 
Compared with \ac{DLS} approaches, \ac{CLS} has the advantages of high convergence quality and native support for non-CNF, i.e., \emph{hybrid} constraints~\cite{kyrillidis2020fouriersat}. 
The main performance bottleneck of \texttt{FourierSAT} and \texttt{GradSAT} is, however, the slow gradient computation on the real domain. \texttt{FourierSAT} needs $O(k^3)$ to compute the gradient  for a constraint of length $k$. 
\texttt{GradSAT} improves this complexity to $O(k^2)$. The computational nature of CLS, especially gradient computation, seems amenable to parallelization, which makes combining \ac{CLS} and \ac{ML}-motivated hardware a promising direction. 
Nevertheless, we reveal that the ideal parallel execution time of \texttt{FourierSAT} and \texttt{GradSAT} is still linear with respect to the length of constraints, i.e., $O^*(k)$, which is hardly satisfactory.

\paragraph{Contributions}
The main theoretical contribution of this paper is a highly-parallel algorithm for computing the \ac{ESPs}, which is the major computation task in \ac{CLS} approaches. 
Given a Boolean formula as the conjunction of Boolean constraints, \texttt{FourierSAT} needs to evaluate the Walsh-Fourier expansions of the constraints.
If the constraints are symmetric, their Walsh-Fourier expansions can be compactly represented by \ac{ESPs}.
Our algorithm is inspired by the \ac{FFT}-based convolution. 
We show that the evaluation of Walsh-Fourier expansions of hybrid Boolean constraints can be vectorized. 
Hence, the gradient computation in \texttt{FourierSAT} can be efficiently implemented by packages frequently used in ML, e.g., JAX~\cite{jax2018github, blondel2022efficient}. 
More specifically, we construct the computation of the Walsh-Fourier expansion as an evaluation trace. 
By reverse traversal of the trace, the gradient can be obtained due to the chain rule, yielding an algorithm for gradient computation with complexity $O(k^2)$, which matches the best known sequential complexity in \texttt{GradSAT}. 
However, our algorithm can be efficiently implemented by a multi-threaded computation paradigm, with sublinear ideal parallel execution time $O^*(\log{k})$ due to the data-level parallelism. 
By further leveraging the instruction-level and thread-level parallelism, the Walsh-Fourier expansions of different constraints can be evaluated on different assignments concurrently. 
The three-level parallelism maps efficiently to the GPU hierarchy of \emph{threads} (data-level), \emph{warps} (instruction-level), and \emph{streaming multiprocessors} (thread-level), enabling ease of GPU acceleration.

Additionally, we propose two optimization techniques to enhance the search efficiency of our approach.
We borrow the idea of \ac{ERWA}~\cite{liang2016exponential} to implement the adaptive constraint reweighting~\cite{cai2020old}, which further exploits the solving history in a parallel search. Furthermore, we apply a rephasing heuristic to balance the search direction between exploitation and exploration.

We implemented our approach as \texttt{FastFourierSAT}, with massively parallel computation on the GPU platform. 
The benchmark instances include a variety of combinatorial optimization problems such as \emph{parity learning} and \emph{weighted MaxCut}. 
\texttt{FastFourierSAT} with GPU implementation computes the gradient 100+ times faster than previous CLS algorithms implemented with CPUs, including \texttt{FourierSAT} and \texttt{GradSAT}. 
We also compare \texttt{FastFourierSAT} with \ac{SOTA} parallel SAT solvers.
Results indicate that \texttt{FastFourierSAT} solves most instances and demonstrates promising scalability.

\section{Preliminaries}
    \subsection{Boolean Constraints and Formulas}
     A Boolean constraint $c$ maps all assignments of a set of $n$ Boolean variables to a truth value, i.e., $c:\mathbb{B}^n\to \{\texttt{True},\texttt{False}\}$. Boolean constraints are often represented by variables and Boolean connectives, e.g., $\wedge$, $\vee$, $\neg$, and $\oplus$.
     
     A Boolean formula $f:\mathbb{B}^n\to \mathbb{B}$ is defined by the conjunction of a set of Boolean constraints $C=\{c_i\}_{i=1}^m$, i.e., $f=c_1\wedge{}c_2\wedge{}\cdots\wedge{}c_m$.  
     A formula is in \ac{CNF} when each constraint is a disjunction of literals. Otherwise, the formula is said to be \emph{hybrid}, which may contain non-\ac{CNF} constraints, e.g., XOR and cardinality constraints~\cite{kyrillidis2020fouriersat}.    

    \subsection{Walsh Expansions of Boolean Constraints}\label{sec:pre_wfe}
        We define a Boolean function by $f:\{\pm{}1\}^n \to \{\pm{}1\}$ (with an overload to $f$) where $-1$ represents \texttt{True} and $+1$ for \texttt{False}. 
        Walsh expansion transforms a Boolean function into a multilinear polynomial, such that the polynomial agrees with the Boolean function on all Boolean assignments~\cite{o2014analysis}. 
        The following theorem shows that every function defined on a Boolean hypercube has an equivalent Walsh expansion.
        
        \begin{theorem}
        (\cite{o2014analysis}, Walsh Transform\footnote{The Walsh transform (expansion), which is also referred to as the Walsh-Fourier transform (expansion) in FourierSAT, is given this specific name to distinguish it from the Fourier transform.}) 
        Given a function $f:\{\pm{}1\}^n \to [-1,1]$, there is a unique way of expressing $f$ as a multilinear polynomial, called the Walsh expansion, with at most $2^n$ terms in $S$, where each term corresponds to one subset of $[n]$ according to:
        \begin{equation}
            \nonumber
            f(x)=\sum_{S\subseteq[n]}\left(
                \hat{f}(S)\cdot\prod_{i\in{}S}x_i
            \right)
        \end{equation}
        where $\hat{f}(S) \in \mathbb{R}^{2^n}$ is called Walsh coefficient, given $S$, and computed as:
        $\hat{f}(S)=\frac{1}{2^n}\sum_{x\in\{\pm{}1\}^n}\left(
            f(x)\cdot\prod_{i\in{}S}x_i
        \right)$
        \label{thm:walshexpansion}
        \end{theorem}
        
        For a Boolean constraint $c$, we use $WE_c$ to denote the Walsh expansion of $c$. 
        \begin{example}
            (Walsh Expansion) 
            Given a cardinality Boolean constraint $c:x_1+x_2+x_3+x_4\ge{}2$, its Walsh expansion is
            \begin{align}
                \nonumber WE_c&(x) = -\frac{3}{8}x_1x_2x_3x_4 \\
                \nonumber -&\frac{1}{8}(x_1x_2x_3+x_1x_2x_4+x_1x_3x_4+x_2x_3x_4)\\
                \nonumber +&\frac{1}{8}(x_1x_2+x_1x_3+x_1x_4+x_2x_3+x_2x_4+x_3x_4)\\
                +&\frac{3}{8}(x_1+x_2+x_3+x_4)-\frac{3}{8}.
                \label{eq:walsh_expansion}
            \end{align}
            \label{ex:walsh_expansion}
        \end{example}

        Note that all literals in the constraint in Eg.~\ref{ex:walsh_expansion} have the same coefficient. 
        Such a constraint is called \emph{symmetric}.
        In the presence of a symmetric constraint, the Walsh coefficient $\hat{f}(S)$ only depends on $|S|$.
        Due to this property, the Walsh expansion of a symmetric constraint can be compactly represented by \ac{ESPs}.
        \begin{definition}
            (Elementary Symmetric Polynomials, ESPs)
            The \ac{ESPs} in $n$ variables $x_1,\cdots,x_n$, denoted by ${esp}(x)=[e_n(x), \cdots, e_0(x)]$, where
            \begin{equation}
                e_j(x)=\sum_{1\le{}i_1<i_2<\cdots<i_j\le{}n}x_{i_1}\cdots{}x_{i_n},\nonumber
            \end{equation}
            \label{def:esp}
            for $1\le j\le n$.
            And $e_0(x)=1$.
        \end{definition}
        
        Therefore, the Walsh expansion in Eg.~\ref{ex:walsh_expansion} can be concisely expressed as the product of two vectors:
        \begin{equation}
            {WE}_c(x) = \hat{f}_c\cdot{}esp(x)^T
            \label{eq:effi_eval}
        \end{equation}
        where $\hat{f}_c\in\mathbb{R}^{n+1}$ is the vector of Fourier coefficients of the constraint $c$, with the $i$-th element as $\hat{f}(S)$, $|S|=i$.
        
    \subsection{Convolution and Fourier Transform}\label{sec:esp_conv}    
        \begin{definition}
            (Linear Convolution)
            With $g\in\mathbb{R}^n$ and $h\in\mathbb{R}^m$ as two one-dimension sequences.
            The linear convolution of $g$ and $h$ is a one-dimension sequence $(g*h)\in\mathbb{R}^{n+m-1}$ 
            Each entry in sequence $(g*h)$ is defined as:
            \begin{equation}
                (g*h)[i]=\sum_{j=0}^{i}g[i-j]\cdot{}h[j].\nonumber
            \end{equation}
        \end{definition}

        \begin{definition}
            (Fourier Transform)
           Let $\omega=\exp{\left(\frac{-2\pi\sqrt{-1}}{n}\right)}$, referred as the base frequency.
           $W\in\mathbb{C}^{n\times{}n}$ is the \ac{DFT} matrix, with each element as $W[i,j]=\omega^{i\cdot{}j}$.
           
           Fourier transform, denoted by $\mathcal{F}(\cdot)$, converts $x\in\mathbb{R}^n$ into $X\in\mathbb{C}^n$ in frequency domain:
            $X=\mathcal{F}(x)=W\cdot{}x$.
            
            Conversely, the inverse Fourier transform is defined by:
            $x=\mathcal{F}^{-1}(X)=W^{-1}\cdot{}X$.
            \label{def:fouriertransform}
        \end{definition}

         Thm.~\ref{thm:conv} states that linear convolution ($*$) in original domain is equivalent to the pointwise multiplication ($\circ$) in frequency domain.
        \begin{theorem}
            (Convolution Theorem)
            With $g$ and $h$ as two one-dimension sequences,
            $g*h = \mathcal{F}^{-1}\left(
                \mathcal{F}(g)\circ\mathcal{F}(h)\right).$
            \label{thm:conv}
        \end{theorem}
        
        Cor. \ref{cor:asso} generalizes Thm. \ref{thm:conv} to multiple sequences.
        \begin{corollary} With $g_1,\cdots, g_k$ as $k$ one-dimension sequences, and $\gamma_i=\mathcal{F}(g_i)$ for $1\le i\le k$, we have
            \begin{equation}
                g_1*g_2*\cdots{}*g_k = \mathcal{F}^{-1}\left(
                \gamma_1\circ\gamma_2\circ\cdots\circ\gamma_k
                \right).
                \label{eq:asso}
            \end{equation}
            \label{cor:asso}
        \end{corollary}

\section{Theoretical Framework}
    In this section, we first recap two CPU implementations of \emph{Continuous Local Search} (\ac{CLS}) for hybrid SAT solving and discuss why they are limited for parallelization.
    We then show how our proposed solver, named \texttt{FastFourierSAT}, can exploit GPU to accelerate gradient computation by parallelizing at \emph{thread}, \emph{instruction}, and \emph{data} levels.

    \texttt{FastFourierSAT} consists of three steps: \emph{i.} Fourier transform, \emph{ii.} multiplications in the frequency domain, and \emph{iii.} inverse Fourier transform.
    
    The complexity of our approach is $O(k^2)$ for computing the gradient of a constraint with length $k$, which matches previous work. 
    Our algorithm is, moreover, highly parallelizable and the ideal execution time with unlimited resources can be reduced to $O^*(\log{k})$.
    
    \subsection{Recap of \ac{CLS}-Based SAT Solving}
        \ac{CLS}-based SAT solvers define a continuous objective function, the minima of which encode the solutions to the original Boolean formula. 
        \begin{definition}
            (Objective) For a formula $f$ with constraint set $C$, the objective function associated with $f$ is defined as:
            \begin{equation}
                F_f(x)=\sum_{c\in{}C}w_c\cdot\texttt{WE}_c(x)
                \label{eq:objective}
            \end{equation}
            where $w_c$ is the weight of $c$ assigned by the \ac{CLS} algorithm.
            \label{def:objective}
        \end{definition}
        
        \begin{theorem}
            (SAT Certificate) Given variables $x\in[-1,1]^n$, a Boolean formula $f$ with constraint set $C$ is satisfiable iff
            $\min{}F_f(x)=-\sum_{c\in{}C}w_c$.
            \label{thm:certificate}
        \end{theorem}

        Based on Def.~\ref{def:objective} and Thm.~\ref{thm:certificate}, previous \ac{CLS} frameworks can be described in Alg.~\ref{alg:cls}, which searches for the ground state of the objective function $F_f$.
        Global optimization on non-convex functions is, however, NP-hard~\cite{jain2017non}.
        Since global optima can be identified efficiently by Thm.~\ref{thm:certificate}, it is usually more practical to converge to local optima and check if any of them is global as in line~5 of Alg.~\ref{alg:cls}.

        \begin{algorithm}[tb]
            \caption{The CLS Framework for Hybrid SAT Solving}
            \label{alg:cls}
            \textbf{Input}: Boolean formula $f$ with a hybrid constraint set $C$\\
            \textbf{Output}: A discrete assignment $x\in\{-1,1\}^n$
            \begin{algorithmic}[1]
                \State Sample $x_0$ uniformly from $[-1,1]^n$
                \State Initialize constraint-weight function $w$
                \For{$j = 1, \cdots, J$}
                    \State $x \gets \texttt{projected-gradient-descent}(F_f,w)$ \Comment{See Appendix~\ref{app:PGD}}
                    \If{$F_f(\texttt{sgn}(x), w)=-\sum_{c\in{}C}w_c$}
                        \State \Return $\texttt{sgn}(x)$
                    \Else 
                        $\text{ }x$, $w$$\gets$\texttt{Restart}($x$, $w$)
                    \EndIf
                \EndFor
                \State \Return $\texttt{sgn}(x)$ with lowest $F_f(\texttt{sgn}(x), w_0)$
            \end{algorithmic}
        \end{algorithm}
        
        \texttt{FourierSAT} and \texttt{GradSAT} \cite{kyrillidis2020fouriersat,kyrillidis2021continuous} are two gradient-based \ac{CLS} variants of  Alg.~\ref{alg:cls} (see Appx.~\ref{app:cls}).
        The majority of the computational workload in \ac{CLS} approaches is line 4 of Alg.~\ref{alg:cls}. Hence, the performance heavily relies on the speed of gradient computation.

        \subsubsection{FourierSAT}
        This approach evaluates the Walsh expansion (Eq.~\ref{eq:effi_eval}) and computes the explicit gradients of each variable.

        \begin{example}
            (Computing the Explicit Gradient)
            For the constraint $x_1+x_2+x_3+x_4\ge{}2$, the partial gradient of Eq.~\ref{eq:walsh_expansion} with respect to $x_1$ is:
            \begin{equation}
                \nonumber{}x_1'=
                \left[\begin{array}{cccc}-\frac{3}{8} & -\frac{1}{8} & \frac{1}{8} & \frac{3}{8}\end{array}\right]
                \left[\begin{array}{c}
                    x_2x_3x_4 \\ x_2x_3+x_2x_4+x_3x_4 \\ x_2+x_3+x_4 \\ 1
                \end{array}\right]
                \label{eq:card}
            \end{equation}
            Given $x=(\frac{1}{2},\frac{1}{2},-\frac{1}{2},-\frac{1}{2})$, $x_1'=\frac{19}{64}$.
            By repeating the above process for computing the partial derivatives with respect to other literals, we will have $x'=(\frac{19}{64},\frac{19}{64},\frac{33}{64},\frac{33}{64})$.
            The gradient guides all variables toward $-1$ to satisfy the constraint.
            \label{ex:fouriersat}
        \end{example}

        Eg.~\ref{ex:fouriersat} demonstrates that differentiating the Walsh expansion with $k$ literals requires computing the partial derivatives for all $k$ variables.
        Computing the partial derivatives for one of the $k$ variables requires $k-2$ convolution operations, with complexity $O(k^2)$. 
        Hence we have the following fact.

        \begin{fact}
            \cite{kyrillidis2020fouriersat}
            \texttt{FourierSAT} computes the gradient via the Walsh expansion. 
            For a symmetric Boolean constraint with $k$ literals, the time complexity of computing the gradient is $O(k^3)$.
        \end{fact}
        
        \subsubsection{GradSAT}
        This approach encodes the Boolean constraint with \ac{BDD}~\cite{bryant1995binary}.
        By performing the belief propagation on  BDDs, the messages are accumulated on the vertices~\cite{pearl1988belief, shafer1990probability}, which can be used to compute the gradient.
        The complexity of this approach depends on the size of the \ac{BDD}~\cite{thornton1994efficient}.
        The \ac{BDD} size of a symmetric Boolean constraint is $O(k^2)$~\cite{sasao1996representations}.

        \begin{fact}
            \cite{kyrillidis2021continuous}
            \texttt{GradSAT} computes the gradient by belief propagation. 
            For a symmetric Boolean constraint with $k$ literals, the time complexity of computing the gradient is $O(k^2)$.
        \end{fact}
       
        \subsubsection{Limited Parallelism of Previous Approaches}
            Consider the ideal parallel execution time\footnote{Consider the execution time as $T=T_s+T_p/N$, where $T_s$, $T_p$ are the serial and parallel components, $N$ is the number of computational resources. We use the $O^*(T)$ notation for $N=\infty$.}, denoted by $O^*(\cdot)$, as the minimum required time for executing an algorithm, given unlimited computational resources.
            \texttt{FourierSAT} needs to convolve the literal sequentially, while \texttt{GradSAT} needs to traverse the \ac{BDD} layer by layer.
            Thereafter, the computation from the preceding literal needs to wait until the computation of the previous literal is completed.
            As a result, the ideal execution time with unlimited resources of the two approaches above is $O^*(k)$ for a constraint with $k$ literals.

    \subsection{FastFourierSAT}
        In this subsection, we describe our approach, named \texttt{FastFourierSAT}.
        We first propose a Fourier-transform-based approach to convert the polynomial-based computation into a vectorized form, which the evaluation trace enables efficient parallelization on GPUs.
        Subsequently, the gradient can be computed by traversing the evaluation trace backward. 
        This approach is also known as \emph{Autodiff}, a fundamental technique that plays a crucial role in \ac{ML}~\cite{baydin2018automatic}.
        We reveal that the complexity of the proposed approach is $O(k^2)$. 
        Moreover, our approach runs in the best theoretical execution time of $O^*(\log{k})$ with a multi-threaded computation scheme.
        
        \subsubsection{Polynomial-based Evaluation}\label{sec:forward}
        The most expensive operations in \texttt{FourierSAT} are the computation of \ac{ESPs}.
        Given a constraint $c$ with a literal set $\{x_1,\cdots,x_k\}$, we observe that the \ac{ESPs} can be obtained by convolutions of $k$ sequences.
        \begin{equation}
            esp(x)=\left[x_1, \; 1\right]*\left[x_2, \; 1\right]*\cdots*\left[x_k, \; 1\right]
            \label{eq:convolution}
        \end{equation}
        By Cor.~\ref{cor:asso}, the convolution operations in Eq.~\ref{eq:convolution} can be computed by 1) pointwise multiplications of sequences in frequency domain; followed by
        2) inverse \ac{FFT} of this result.

        \begin{algorithm}[tb]
            \caption{Forward Evaluation with FFT method}
            \label{alg:forward}
            \textbf{Input}: The current assignment $x$\\
            \textbf{Parameter}: Conjugated Fourier coefficient $\tilde{f}_c$'s $\forall{}c\in{}C$\\
            \textbf{Output}: Value $F_f$
            \begin{algorithmic}[1]
                \State Initialize $F_f=0$
                \For{$c\in{}C$, \textbf{parallel}}          \Comment{Instruction-level parallelism}
                    \State $\Gamma_c=\left[1 \ \omega \ \cdots \ \omega^{\text{len}(c)}\right]^T+x[c]$         \Comment{Eq.~\ref{eq:forward}, step~i\;\;}
                    \State $\gamma_c=\texttt{prod}(\Gamma_c, \text{axis}=1)$ \Comment{Eq.~\ref{eq:reduction}, step~ii\;}
                    \State $\text{WE}_c=\tilde{f}_c\gamma_c$     \Comment{Eq.~\ref{eq:inverse}, step~iii}
                    \State $\texttt{atomicAdd}(\&F_f, \text{WE}_c)$
                \EndFor
                \State \Return $F_f$ 
            \end{algorithmic}
        \end{algorithm}

        In the following, we describe the details of the three main steps of \texttt{FastFourierSAT} as Lines 3-5 in~Alg.~\ref{alg:forward}.
        \paragraph{i. Fourier Transform as Addition:}
            Given a Boolean constraint $c$ with $k$ literals, $esp(x)$ is a one-dimension sequence with $k+1$ entries.
            Prior to performing batched Fourier transform on sequences $[x_i,\;1]$ for all $1\le i\le k$, it is necessary to extend them with 0's, resulting in a sequence with a length of $k+1$, i.e., $g_i=[x_i,\;1,\;0,\;\cdots,\;0]$.
            By Def.~\ref{def:fouriertransform}, the sequences in the frequency domain can be obtained by:
            \begin{equation}
                \nonumber\gamma_i^T=W\cdot{}g_i^T=\left[\begin{array}{cccc}
                1+x_i & \omega+x_i & \cdots & \omega^k+x_i\end{array}\right]^T.
            \end{equation}
            Hence, by eliminating the trivial operations, the batched Fourier transform of $k$ sequences can be described as an outer addition of two vectors:
            \begin{equation}
                \left[\gamma_1 \      \gamma_2 \  \cdots \    \gamma_k\right]
                =\left[1 \      \omega \    \cdots \    \omega^k\right]^T
                +\left[x_1 \    x_2 \       \cdots \    x_k\right].
                \label{eq:forward}
            \end{equation}

        \paragraph{ii. Multiplication in frequency domain:}
            By performing row-wise reduce product for the left-hand-side of Eq.~\ref{eq:forward}, the elements encoding different variables but with the same frequencies are accumulated.
            The resulting column vector $\gamma_{1:k}$ is the \ac{ESPs} in frequency domain of due to Cor.~\ref{cor:asso}.
            \begin{align}
                \label{eq:reduction}\gamma_{1:k} &= \gamma_1\circ\gamma_2\circ\cdots\circ\gamma_k\\ 
                \nonumber&= \left[
                    \prod_{i=1}^k\left(1+x_i\right) \ \prod_{i=1}^k(\omega+x_i) \ \cdots \ \prod_{i=1}^k(\omega^k+x_i)
                \right]^T
            \end{align}

        \paragraph{iii. Inverse Transform as Vector Multiplication:} 
            By applying the inverse Fourier transform, $\gamma_{1:k}$ can be converted back to the \ac{ESPs}.
            The Walsh expansion is then obtained by multiplying the Fourier coefficients with the \ac{ESPs}.
            Thereafter, Eq.~\ref{eq:effi_eval} can be written as:
            \begin{equation}
                WE_c(x) = 
                    \lefteqn{\overbrace{\phantom{\hat{f}_c\cdot{}W^{-1}}}^{\tilde{f}_c}}\hat{f}_c\cdot{}
                    \underbrace{W^{-1}\cdot\gamma_{1:k}(x)}_{esp(x)}.
                \label{eq:inverse}
            \end{equation}
            
            However, given a constraint $c$, $\hat{f}_c$ and $W^{-1}$ are fixed during the computation time. 
            So $\tilde{f}_c\in\mathbb{C}^{k+1}=\hat{f}_c\cdot{}W^{-1}$ can be computed in preprocessing, and we denote the resulting row vector as conjugated Walsh coefficient.

        In the following, we give an example to demonstrate how the three steps of \texttt{FastFourierSAT} work.
        
        \begin{example}
            (Fourier Transform-based Evaluation)\label{ex:forward}
            Given a constraint $x_1+x_2+x_3+x_4\ge{}2$, and base frequency $\omega=\exp\left(\frac{-2\pi\sqrt{-1}}{5}\right)$, the conjugated Walsh coefficient is
            \begin{equation}
                \nonumber
                \tilde{f}_c^T=-\frac{1}{40}\left[\begin{array}{c}
                    3 \\ 
                    \omega^4-\omega^3-3\omega^2+3\omega+3 \\ 
                    -3\omega^4+\omega^3+3\omega^2-\omega+3 \\
                    -\omega^4+3\omega^3+\omega^2-3\omega+3 \\ 
                    3\omega^4-3\omega^3-\omega^2+\omega+3
                \end{array}\right]
            \end{equation}
            And the evaluation trace will be constructed as:
            \begin{center}
                \tiny
                \begin{tikzpicture}[round/.style={circle, draw=black!100, thick, minimum size=6.5mm},
                trans/.style={rectangle, draw=black!0, thick, minimum width=1mm, minimum height=1mm},]
                    \newcommand{\y}{0.4}
                    \newcommand{\x}{1.35}
                    \node[round] (x1) at (0, 0) {$x_1$};
                    \node[round] (x2) at (0, -2*\y) {$x_2$};
                    \node[round] (x3) at (0, -4*\y) {$x_3$};
                    \node[round] (x4) at (0, -6*\y) {$x_4$};
    
                    \node[round] (y1) at (\x, 0) {$\gamma_1$};
                    \node[round] (y2) at (\x, -2*\y) {$\gamma_2$};
                    \node[round] (y3) at (\x, -4*\y) {$\gamma_3$};
                    \node[round] (y4) at (\x, -6*\y) {$\gamma_4$};
    
                    \node[round] (y12) at (2*\x, -1*\y) {\tiny $\gamma_{1:2}$};
                    \node[round] (y34) at (2*\x, -5*\y) {\tiny $\gamma_{3:4}$};
    
                    \node[round] (yc) at (3*\x, -3*\y) {$\gamma_{1:4}$};
                    \node[trans] (we) at (4*\x, -3*\y) {$WE_c$};
                    
                    \newcommand{\arrowIn}{\tikz \draw[-stealth] (-1pt,0) -- (1pt,0);}
    
                    \draw[thick] (x1.east) -- (y1.west) node[sloped,pos=0.5,allow upside down]{\arrowIn};
                    \draw[thick] (x2.east) -- (y2.west) node[sloped,pos=0.5,allow upside down]{\arrowIn};
                    \draw[thick] (x3.east) -- (y3.west) node[sloped,pos=0.5,allow upside down]{\arrowIn};
                    \draw[thick] (x4.east) -- (y4.west) node[sloped,pos=0.5,allow upside down]{\arrowIn};
                    \draw[thick] (y1.east) -- (y12.north west) node[sloped,pos=0.5,allow upside down]{\arrowIn};
                    \draw[thick] (y2.east) -- (y12.south west) node[sloped,pos=0.5,allow upside down]{\arrowIn};
                    \draw[thick] (y3.east) -- (y34.north west) node[sloped,pos=0.5,allow upside down]{\arrowIn};
                    \draw[thick] (y4.east) -- (y34.south west) node[sloped,pos=0.5,allow upside down]{\arrowIn};
                    \draw[thick] (y12.east) -- (yc.north west) node[sloped,pos=0.5,allow upside down]{\arrowIn};
                    \draw[thick] (y34.east) -- (yc.south west) node[sloped,pos=0.5,allow upside down]{\arrowIn};
                    \draw[thick] (yc.east) -- (we.west) node[sloped,pos=0.5,allow upside down]{\arrowIn};
                \end{tikzpicture}
            \end{center}

            i. Consider $x=\left(\frac{1}{2},\frac{1}{2},-\frac{1}{2},-\frac{1}{2}\right)$, we construct the corresponding column vector $\left[\frac{1}{2},\frac{1}{2},-\frac{1}{2},-\frac{1}{2}\right]^T$ for the batched Fourier transform.
            By performing outer addition with $[1,\omega,\omega^2,\omega^3,\omega^4]$, the sequences in frequency domain are:
            \begin{align}
                \nonumber\gamma_1=\gamma_2&=[\frac{3}{2}, \omega+\frac{1}{2}, \omega^2+\frac{1}{2}, \omega^3+\frac{1}{2}, \omega^4+\frac{1}{2}]^T\\
                \nonumber\gamma_3=\gamma_4&=[\frac{1}{2}, \omega-\frac{1}{2}, \omega^2-\frac{1}{2}, \omega^3-\frac{1}{2}, \omega^4-\frac{1}{2}]^T
            \end{align}
            
            ii. By multiplying in the frequency domain along the binary tree, we can have:
            \begin{align}
                \nonumber\gamma_{1:2}&=\gamma_1\circ\gamma_2=[\frac{3}{2}, (\omega+\frac{1}{2})^2, \cdots, (\omega^4+\frac{1}{2})^2]^T\\
                \nonumber\gamma_{3:4}&=\gamma_3\circ\gamma_4=[\frac{1}{2}, (\omega-\frac{1}{2})^2, \cdots, (\omega^4-\frac{1}{2})^2]^T\\
                \nonumber\gamma_{1:4}&=\gamma_{1:2}\circ\gamma_{3:4}=[0, (\omega^2-\frac{1}{4})^2, \cdots, (\omega^8-\frac{1}{4})^2]^T
            \end{align}
            
            iii. As in Eq.~\ref{eq:inverse}, the Walsh expansion of $c$ can be evaluated as $WE_c(x)=-\frac{263}{640}+\frac{\omega+\omega^2+\omega^3+\omega^4}{20}=-\frac{59}{128}$.
        \end{example}

        \begin{theorem}
            (Forward Evaluation)
            Lines 3-5 in Alg.~\ref{alg:forward} compute the Walsh expansion based on FFT. 
            For a symmetric Boolean constraint with $k$ literals, this algorithm runs in $O(k^2)$ time.
            \label{thm:forward}
        \end{theorem}

        \subsubsection{Graph-based Differentiation}            
            Given a function $f$, a computation graph can be constructed with the input and output nodes corresponding to the variables $x$ and the value $WE(x)$.
            In the forward phase, the function is computed forward with the original operator and recording the intermediate variables $\gamma_i$.
            In the backward phase, derivatives are calculated by the differential operators, and local gradients $\gamma_i'$ are propagated in reverse.
            The differential operators are derived based on:
            
            \begin{proposition}
                (Chain rule) Given $x$ and composite functions $WE_c\left(\gamma_c(x)\right)$, which is differentiable, then we have
                \begin{equation*}
                    \left.\frac{\partial{WE}_c}{\partial{}x_i}\right|_x
                    =\left.\frac{\partial{WE}_c}{\partial\gamma_c}\right|_{\gamma_c(x)}
                    \left.\cdot\frac{\partial\gamma_c}{\partial{}x_i}\right|_x.
                \end{equation*}
                \label{prop:chainrule}
            \end{proposition}
            In the backward phase, the gradients can be computed by traversing the computation graph from top to bottom.

            \begin{example}
                (Autodiff-based Differentiation)\label{ex:backward}
                Given a constraint $x_1+x_2+x_3+x_4\ge{}2$, from Prop.~\ref{prop:chainrule} we can derive:
                \begin{align}
                    \nonumber{}x_1'=\mathds{1}\cdot&\left(\left(\tilde{f}_c\circ\gamma_{3:4}\right)\circ\gamma_2\right)\\
                    &\vdots\label{eq:autodiff}\\
                    \nonumber{}x_4'=\mathds{1}\cdot&\left(\left(\tilde{f}_c\circ\gamma_{1:2}\right)\circ\gamma_3\right)
                \end{align}
                where $\mathds{1}$ is a row vector with all elements as 1 and $\circ$ denotes the point-wise multiplication of two column vectors.
                Hence, the trace in backward traversal will be:
                \begin{center}
                    \tiny
                    \begin{tikzpicture}[round/.style={circle, draw=black!100, thick, minimum size=6.5mm},
                    trans/.style={rectangle, draw=black!0, thick, minimum width=1mm, minimum height=1mm},]
                        \newcommand{\y}{0.4}
                        \newcommand{\x}{1.35}
                        \node[round] (x1) at (0, 0) {\scriptsize$x_1'$};
                        \node[round] (x2) at (0, -2*\y) {\scriptsize$x_2'$};
                        \node[round] (x3) at (0, -4*\y) {\scriptsize$x_3'$};
                        \node[round] (x4) at (0, -6*\y) {\scriptsize$x_4'$};
        
                        \node[round] (y1) at (\x, 0) {$\gamma_2$};
                        \node[round] (y2) at (\x, -2*\y) {$\gamma_1$};
                        \node[round] (y3) at (\x, -4*\y) {$\gamma_4$};
                        \node[round] (y4) at (\x, -6*\y) {$\gamma_3$};
        
                        \node[round] (y12) at (2*\x, -1*\y) {\tiny $\gamma_{3:4}$};
                        \node[round] (y34) at (2*\x, -5*\y) {\tiny $\gamma_{1:2}$};
        
                        \node[round] (yc) at (3*\x, -3*\y) {$\tilde{f}_c$};
                        \node[trans] (we) at (4*\x, -3*\y) {$WE_c$};
                        
                        \newcommand{\arrowIn}{\tikz \draw[-stealth] (-1pt,0) -- (1pt,0);}
        
                        \draw[thick] (y1.west) -- (x1.east) node[sloped,pos=0.5,allow upside down]{\arrowIn};
                        \draw[thick] (y2.west) -- (x2.east) node[sloped,pos=0.5,allow upside down]{\arrowIn};
                        \draw[thick] (y3.west) -- (x3.east) node[sloped,pos=0.5,allow upside down]{\arrowIn};
                        \draw[thick] (y4.west) -- (x4.east) node[sloped,pos=0.5,allow upside down]{\arrowIn};
                        \draw[thick] (y12.north west) -- (y1.east) node[sloped,pos=0.5,allow upside down]{\arrowIn};
                        \draw[thick] (y12.south west) -- (y2.east) node[sloped,pos=0.5,allow upside down]{\arrowIn};
                        \draw[thick] (y34.north west) -- (y3.east) node[sloped,pos=0.5,allow upside down]{\arrowIn};
                        \draw[thick] (y34.south west) -- (y4.east) node[sloped,pos=0.5,allow upside down]{\arrowIn};
                        \draw[thick] (yc.north west) -- (y12.east) node[sloped,pos=0.5,allow upside down]{\arrowIn};
                        \draw[thick] (yc.south west) -- (y34.east) node[sloped,pos=0.5,allow upside down]{\arrowIn};
                        \draw[thick, densely dotted] (we.west) -- (yc.east) node[sloped,pos=0.5,allow upside down]{\arrowIn};
                    \end{tikzpicture}
                \end{center}
                
                In the evaluation stage (Eg.~\ref{ex:fouriersat}), \texttt{FastFourierSAT} has recorded $\tilde{f}_c$ and $\gamma_i$'s at $x=(\frac{1}{2},\frac{1}{2},-\frac{1}{2},-\frac{1}{2})$.
                By traversing the computation graph in backward, i.e., solving Eq.~\ref{eq:autodiff} from inner parentheses to outside, the resulting gradient $x'=(\frac{19}{64},\frac{19}{64},\frac{33}{64},\frac{33}{64})$ matches Eg.~\ref{ex:fouriersat}.
            \end{example}
        
            \begin{theorem}
                (Backward Differentiation)
                The differential operators derived from Prop.~\ref{prop:chainrule} can compute the local gradients with the intermediate variables recorded in Alg.~\ref{alg:forward}. 
                It runs in $O(k^2)$ time for a symmetric Boolean constraint with $k$ literals. 
                \label{thm:backward}
            \end{theorem}


            

        \subsubsection{Acceleration with GPU}\label{sec:gpu}
            The majority of the computations in Alg.~\ref{alg:forward} are in Lines 3-5.
            With the multi-threaded computation scheme, the execution can be parallelized.
            The details are included in the proof of the following proposition.
            \begin{proposition}
                (Parallelism) With unbounded computational resources, the ideal execution time with unlimited resources of Autodiff for Alg.~\ref{alg:forward} scales at $O^*(\log{k})$.
                \label{prop:parallel}
            \end{proposition}
        
        Parallel \ac{SAT} solving refers to the process of solving a \ac{SAT} formula using multiple computational resources simultaneously.
        It is motivated by the desire to exploit the computational power of modern multi-core processors to speed up the solving process.
        We leverage massive parallelism to achieve the ideal execution time in Prop.~\ref{prop:parallel}.
        \paragraph{i. Data-level Parallelism ($p_d$)}
            The majority of computations of \texttt{FastFourierSAT} are matrix operations, which are highly optimized for GPUs.
            The elementary operations are mainly additions and multiplications, which are simple instructions.
            Due to the \emph{single instruction multiple data} scheme, concurrent execution of these simple instructions can be efficiently handled by a warp with a group of \emph{threads}.
        \paragraph{ii. Instruction-level Parallelism ($p_i$)}
            The differentiation of the objective function (Eq.~\ref{eq:objective}) can be subdivided into the differentiation of Walsh expansions (Eq.~\ref{eq:effi_eval}) of all Boolean constraints.
            Since these computations are data independent, they can be instruction-level parallelized, and distributed to identical \emph{warps} in a streaming multiprocessor.
        \paragraph{iii. Thread-level Parallelism ($p_t$)} 
            In \ac{CLS} solvers, different initialization can be assigned to identical optimizers for parallel search.
            The search trajectory becomes diversified as optimizers might converge to different local optima.
            It increases the chance of finding a local optimum being globally optimal.
            The parallel search can be partitioned and solved by many \emph{streaming multiprocessors}. 

        
            
\section{Implementation Techniques}
        The essence of \ac{CLS}-based SAT solving is using continuous optimization to find the ground states of the non-convex energy landscape.
        The optimization results depend heavily on the initialization~\cite{jain2013low}.
        To search in the continuous domain in a more systematic way, we incorporate heuristics with the algorithm upon restart.
        The ablation study and illustrative examples are  in Appx.~\ref{app:heuristic}.

        \paragraph{Weighting Heuristics}
            To leverage the information along the search history, we propose to model the adaptive weight with \emph{exponential recency weighted average} (ERWA), a lightweight method for progressively approximating a moving average~\cite{liang2016exponential}.
            For each constraint $c$, the weighting heuristic maintains an integer $0 \le{}U_c \le{}p_t$, which is the number of tasks where $c$ appears to be unsatisfied among the $p_t$ parallel tasks.
            We use the normalized $r_c$ to describe the unsatisfaction score: $r_c = U_c/\max_{c\in{}C_f}(U_c)$. If a constraint $c$ is more on the satisfied (resp. unsatisfied) side, $r_{c}$ is closer to 0 (resp. 1).
            
            Consider the unsatisfaction scores along the search history $r^{(1)}_c,\cdots,r^{(t)}_c$.
            We model the adaptive weight $w_c$ using the \ac{ERWA} of $r^{(i)}_c$'s, i.e, $w^{(t)}_c = \sum_{i=1}^t\beta_ir^{(i)}_c$, with decay rate $\beta_i=\alpha(1-\alpha)^{t-i}$ and $\alpha\in[0,1]$.
            A constraint adapts to a higher weight if it is frequently unsatisfied along the search history. 
            Therefore, the solver should focus more on the subspace that satisfies constraints with a higher $w_c$. 
            In practice, the adaptive weight can be efficiently updated as follows.
            \begin{proposition}
                (Adaptive weight)
                The weight of $c$ at step $t+1$ can be computed by  $ w^{(t+1)}_c = (1-\alpha)w^{(t)}_c+\alpha{}r_c^{(t)}$.
                \label{prop:erwa}
            \end{proposition}

        \paragraph{Rephasing Heuristics}\label{sec:rephase}
            The aim of rephasing is to balance between focusing the search on the neighboring subspace (intensification) and searching in an entirely different subspace (diversification).
            To this end, we propose a rephasing heuristic that will switch between the following phases:
            \begin{itemize}
                \item \emph{\underline{O}riginal phase}: a point taken from the optimization result of the previous iteration.
                \item \emph{\underline{F}lipped phase}: a point that each variable takes the opposite value to the original phase.
                \item \emph{\underline{R}andom phase}: a point randomly sampled from $[-1,1]^n$, where $n$ is the number of variables.
            \end{itemize}
            \texttt{FastFourierSAT} exploits the original search space in the original phase whereas it explores different search spaces in the flipped phase and random phase.
            To balance between exploitation and exploration, we describe policies of rephasing in the Appx.~\ref{app:imple}.
            
\section{Experimental Results}
    \begin{figure}[tb]
        \centerline{\includegraphics[width=0.5\textwidth]{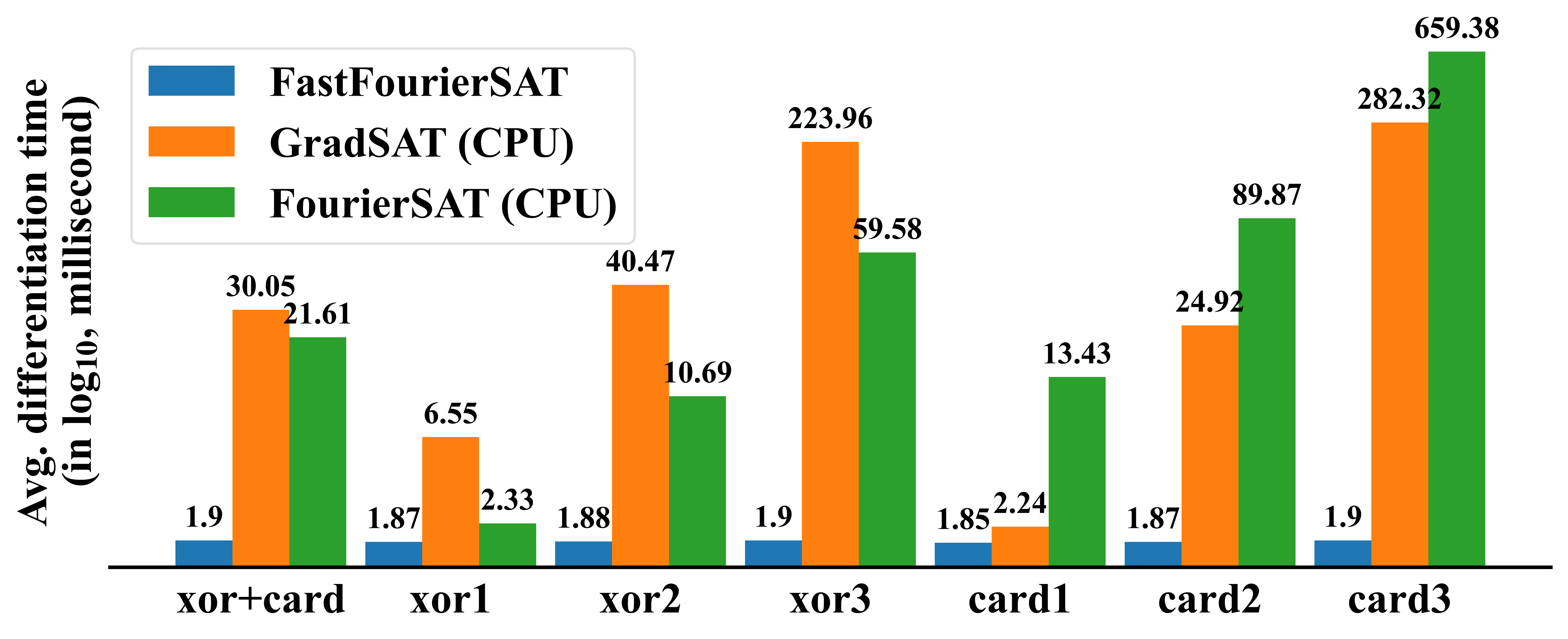}}
        \caption{The average running time of the gradient computation using different \ac{CLS} approaches.}
        \label{fig:grad}
    \end{figure}

    In this section, we compare our solver \texttt{FastFourierSAT} with \ac{CLS} and other \ac{SOTA} SAT solvers.
    We aim to conduct experiments to answer the following research questions:
    
    \textbf{RQ1.} Can implementing \texttt{FastFourierSAT} on GPUs leads to a substantial acceleration in gradient computations compared to prior CPU implementations of \ac{CLS}?

    \textbf{RQ2.} Can \texttt{FastFourierSAT} outperform \ac{SOTA} parallel solvers in solving satisfiability problems that can be naturally encoded with hybrid Boolean constraints?
    
    \textbf{RQ3.} What is the advantage of \texttt{FastFourierSAT} on solving discrete optimization problems such as \emph{Max-Cut}?

    For RQ1, we randomly generate Boolean formulas consisting of XORs and cardinalities.
    The gradient computation speed is compared across the \ac{CLS} algorithms.
    For RQ2 and RQ3, we generate three classes of Boolean formulas that naturally encode: \emph{i.} \emph{random cardinality constraints}, \emph{ii.} \emph{parity learning with error}, and \emph{iii.} \emph{weighted Max-Cut} problems.
    We compare our approach with \emph{a.} solvers in SAT competitions, \emph{b.} GPU-accelerated solvers, and \emph{c.} solvers in Max-SAT Evaluations.
    The details can be found in Appx.~\ref{app:benchmark}.

    \begin{table*}[tb]
        \addtolength{\tabcolsep}{-3pt}
        \centering
        \begin{tabular}{ccccccccccccccccccccc}
        \hline
         &  & \multicolumn{9}{c}{Average Score (\%)} &  & \multicolumn{9}{c}{Number of Best Solutions Achieved} \\ \cline{3-11} \cline{13-21} 
        \multirow{-2}{*}{Methods} &  & MSE19 &  & 16, 16 &  & 16, 32 &  & 32, 16 &  & 32, 32 &  & MSE19 &  & 16, 16 &  & 16, 32 &  & 32, 16 &  & 32, 32 \\ \cline{1-1} \cline{3-3} \cline{5-5} \cline{7-7} \cline{9-9} \cline{11-11} \cline{13-13} \cline{15-15} \cline{17-17} \cline{19-19} \cline{21-21} 
        \textbf{FastFourierSAT (CLS)} &  & \textbf{100.00} &  & \textbf{100.00} &  & \textbf{99.90} &  & \textbf{99.96} &  & \textbf{99.71} &  & \textbf{102} &  & \textbf{82} &  & \textbf{82} &  & \textbf{90} &  & \textbf{69} \\
        VBS w/o CLS &  & 99.13 &  & 99.94 &  & 99.23 &  & 99.15 &  & 99.43 &  & 101 &  & 21 &  & 21 &  & 10 &  & 31 \\
        NuWLS (DLS) &  & 99.13 &  & 99.88 &  & 98.94 &  & 98.79 &  & 98.82 &  & 101 &  & 13 &  & 13 &  & 4 &  & 14 \\
        NuWLS-c (CDCL+DLS) &  & 99.13 &  & 99.6 &  & 98.11 &  & 98.28 &  & 98.29 &  & 101 &  & 4 &  & 4 &  & 5 &  & 5 \\
        SATLike (DLS) &  & 99.09 &  & 97.33 &  & 95.78 &  & 95.63 &  & 97.22 &  & 101 &  & 3 &  & 3 &  & 1 &  & 4 \\
        SATLike-ck (CDCL+DLS)&  & 99.09 &  & 96.04 &  & 95.36 &  & 95.57 &  & 97.47 &  & 100 &  & 1 &  & 1 &  & 0 &  & 8 \\
        GradSAT (CLS) &  & 99.13 &  & 95.90 &  & 94.51 &  & 94.68 &  & 0.00~* &  & 101 &  & 0 &  & 0 &  & 0 &  & 0 \\
        TT-Open-WBO (CDCL) &  & 10.38 &  & 0.09 &  & 0.03 &  & 0.03 &  & 0.00 &  & 9 &  & 0 &  & 0 &  & 0 &  & 0 \\
        \hline
        \end{tabular}
        \caption{Results on weighted Max-Cut problems. * means the solver can not give a valid solution within the given time limit.}
        \label{tab:maxcut}
    \end{table*}

    \textbf{RQ1.} 
        The comparison of average gradient computation time using different \ac{CLS} approaches is shown in Fig.~\ref{fig:grad}.
        For small-scale formulas, \texttt{FastFourierSAT} demonstrates a speedup of approximately $1.2\times$ on \texttt{xor1} and \texttt{card1}. 
        However, when differentiating larger formulas, this speedup becomes even more substantial, reaching $31.36\times$ on \texttt{xor3} and an impressive $148.59\times$ on \texttt{card3}.
        In fact, the majority of the additional overhead is experienced in the preprocessing stage.
        After that, GPU and CPU do not need to frequently access the memory from each other.
        After that initial cost, the benefits of GPU acceleration far outweigh the costs of data movement, resulting in a significant speedup over existing approaches for large instances.

    \begin{figure}[tb]
        \centerline{\includegraphics[width=0.5\textwidth]{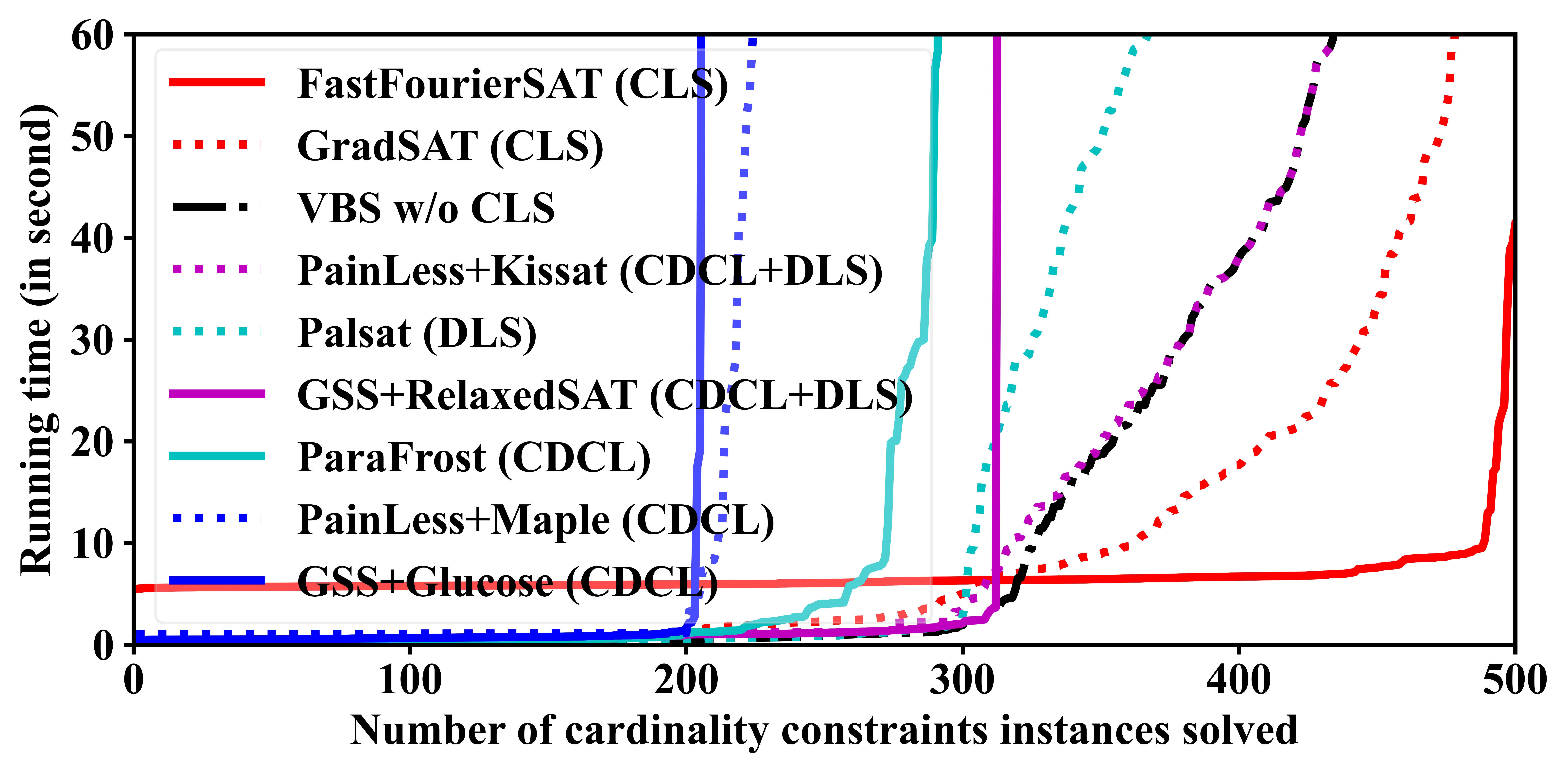}}
        \caption{Results on random cardinality constraints. Solid lines are referred to GPU solvers.}
        \label{fig:card}
    \end{figure}
    
    \textbf{RQ2.}
        The results of Benchmark 1 are shown in Fig.~\ref{fig:card}.
        The largest instance in this benchmark will encode into a \ac{CNF} formula with more than 90000 variables and 180000 clauses.
        In contrast, the cardinality constraints can be natively accepted by \texttt{GradSAT} and \texttt{FastFourierSAT}.
        \texttt{GradSAT} (478) solved more problems than the virtual best solver without \ac{CLS} (\texttt{VBS w/o CLS}) (434).
        \texttt{FastFourierSAT} can solve all the instances as it is encoding-free and the computations can be massively parallelized on GPUs.

\begin{figure}[tb]
        \centerline{\includegraphics[width=0.5\textwidth]{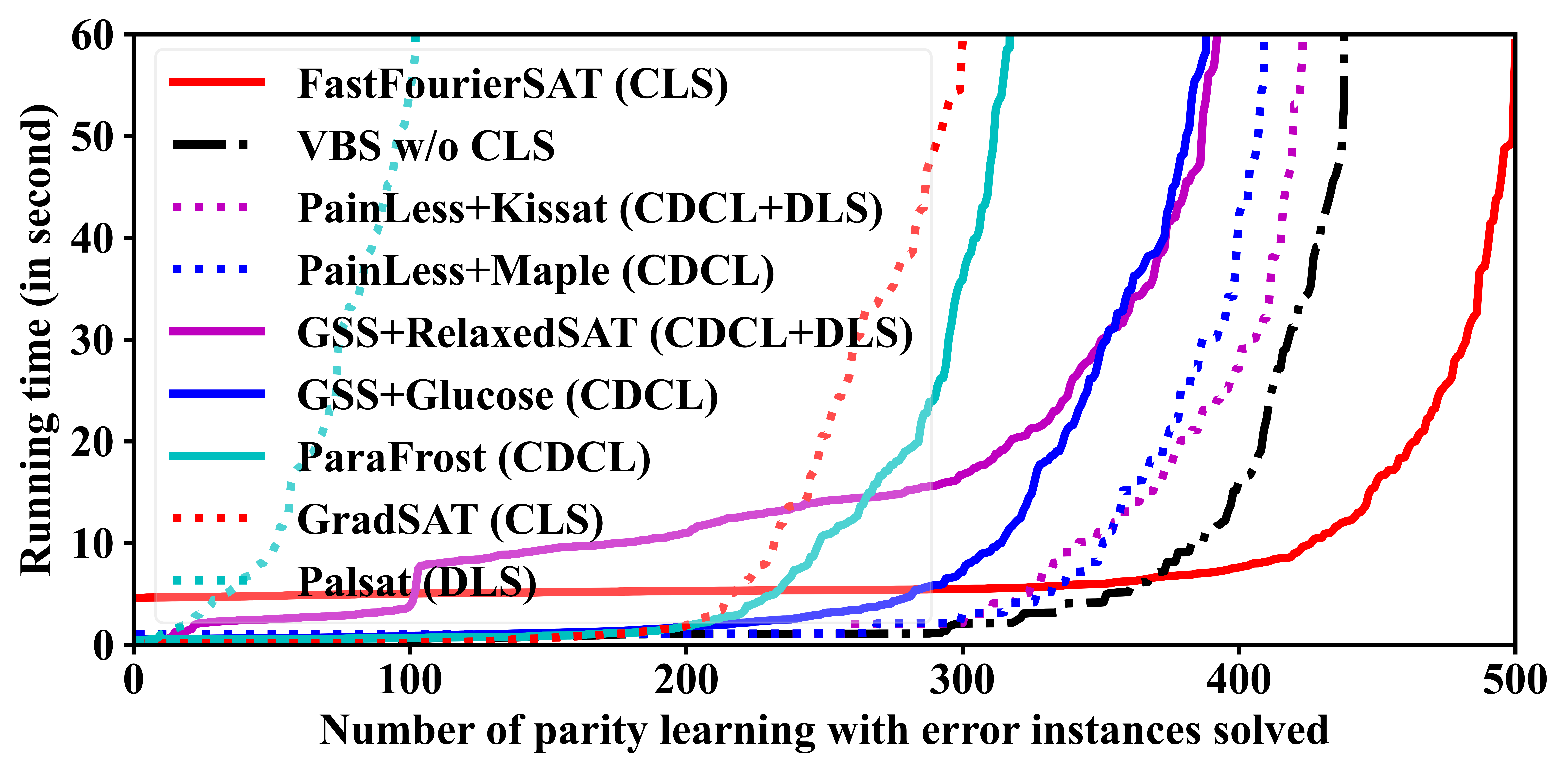}}
        \caption{Results on parity learning with error problems. Solid lines are referred to GPU solvers.}
        \label{fig:parity}
    \end{figure}

        The results of Benchmark 2 are shown in Fig.~\ref{fig:parity}.
        \texttt{GradSAT} (300) solved fewer problems than \texttt{VBS w/o CLS} (438).
        On one hand, the problem of this benchmark is known to be less challenging for the solvers with systematic search (i.e., \ac{CDCL}) than local search~\cite{hoos1999systematic}, where \texttt{PalSAT} can solve 102 problems only.
        On the other hand, the encoding is less expensive than the previous benchmarks, where the largest \ac{CNF} encoding has around 6500 variables and 20000 clauses.
        With the proposed implementation techniques, \texttt{FastFourierSAT} can diversify the search trajectories to many subspaces, which increases the probability of finding a solution.
        Aided by the GPU accelerated gradient computation, \texttt{FastFourierSAT} can solve all the instances in this benchmark.

    \textbf{RQ3.}
        The results of Benchmark 3 are shown in Table~\ref{tab:maxcut}.
        In this benchmark, the local search solvers can achieve better solutions than the \ac{CDCL} solver.
        In the ``MSE19'' category, most solvers, with the exception of \texttt{TT-Open-WBO}, can find the best known solutions.
        This category showcases the effectiveness of these solvers in solving small graphs with practical and theoretical interest~\cite{johnson1996cliques}.
        
        The application of the weighted Max-Cut problem in a planted partition graph is useful in the community detection of social networks~\cite{xu2007scan}.
        We generated planted partition graphs across various scales, in which the largest instance consists of 32 communities, each comprising 32 vertices, yielding a graph with more than $260000$ edges.
        This instance size makes it insufficient for \texttt{GradSAT} to complete one iteration of Alg.~\ref{alg:cls} within 60 second.
        \texttt{VBS w/o CLS} can find the best solution of $83$ instances from this categories.
        \texttt{FastFourierSAT} surpasses this performance by successfully finding the best solution for $323$ instances.
        The results indicate that massive parallelism with GPU can be powerful in generating high-quality solutions for large-scale combinatorial optimization problems such as Max-Cut.

\section{Conclusion}
    The present study introduces a novel \ac{FFT}-based approach for accelerating the evaluation of \ac{ESPs}, which significantly enhances the \ac{CLS} approach for hybrid SAT solving.
    The acceleration is achieved by leveraging the thread-level, instruction-level, and data-level parallelism on the GPU, which enables efficient gradient computation as compared to the prototype of our proposed approach.
    Furthermore, we proposed restart heuristics in the parallel search to mitigate the effects of local optima and saddle points.
    Our results demonstrate that \texttt{FastFourierSAT} is competitive with the recent competition-winning \ac{SOTA} solvers in the three benchmarks.
    They also highlight that, with the massively parallel scheme, \ac{CLS} approaches can complement the existing solvers for solving non-\ac{CNF} constraints.

    It should be noted, however, that in our proposed FFT-based approach, high bit precision is necessary to alleviate the errors introduced by the numerical round-off. 
    Consequently, the overhead in both computation and memory usage may be significant. 
    In the future, it is worthwhile to explore approaches that can lower the required bit precision while bounding the error to within a reasonable value.

\section*{Acknowledgements}
We thank Moshe Vardi for the helpful and insightful comments and discussions that contributed to this work.
This work is funded in part by the  Agency for Science, Technology and Research (A*STAR), Singapore, under its Programmatic Grant Programme (A18A6b0057) and by the National Research Foundation (NRF), Singapore, under its the Competitive Research Programme (CRP Award No. NRF-CRP24-2020-0003).
Zhiwei Zhang is supported in part by NSF grants (IIS-1527668, CCF1704883, IIS-1830549), DoD MURI grant (N00014-20-1-2787), Andrew Ladd Graduate Fellowship of Rice Ken Kennedy Institute, and an award from the Maryland Procurement Office.
The computational work for this article was performed on resources of the National Supercomputing Centre, Singapore (https://www.nscc.sg).
\bigskip

\bibliography{aaai24}

\newpage
\appendix
\onecolumn
\section{Continuous Local Search Algorithms}\label{app:cls}
    \subsection{FourierSAT}
        If the Walsh expansion is a $k$-th order polynomial, then the explicit form of the partial derivative will be a ($k-1$)-th order polynomial, in which the main operations are computing the \ac{ESPs} of $k-1$ variables.
        
        \begin{algorithm}[htbp]
            \caption{Differentiation with Explicit Form Gradient}
            \label{alg:fouriersat}
            \textbf{Input}: Boolean formula $f$ with constraint set $C$ and the current assignment $x$\\
            \textbf{Parameter}: Fourier coefficient $\hat{f}$\\
            \textbf{Output}: Gradient $\nabla_xF_f$
            \begin{algorithmic}[1]
                \State Initialize gradient $\nabla_xF_f=[0\;0\;\cdots\;0]$
                \For{$c\in{}C$}
                    \For{$l\in{}c$}
                        \State Initialize empty sequence $\gamma_c$
                        \For{$l'\in{}c\setminus{}l$}
                            \State $\gamma_c=$\texttt{convolve}($\gamma_c$, [1\;$x_{l'}$])
                        \EndFor
                        \State $\partial_{x_l}F_f+=\hat{f}_c[:-1]\cdot{}\gamma_c$    \Comment{The constant term (last term) of the Walsh expansion becomes 0 in derivatives}
                    \EndFor
                \EndFor
                \State \Return $\nabla_xF_f$ 
            \end{algorithmic}
        \end{algorithm}

        Eg.~\ref{ex:fouriersat} is a representative example that showcases the gradient computation in \texttt{FourierSAT}.
        Given $x=(\frac{1}{2},\frac{1}{2},-\frac{1}{2},-\frac{1}{2})$, the partial derivatives of other variables can be derived as:
        \begin{align*}
            &x_1'= -\frac{3}{8}x_2x_3x_4 -\frac{1}{8}(x_2x_3+x_2x_4+x_3x_4) +\frac{1}{8}(x_2+x_3+x_4) +\frac{3}{8}=\frac{19}{64}\\
            &x_2'= -\frac{3}{8}x_1x_3x_4 -\frac{1}{8}(x_1x_3+x_1x_4+x_3x_4) +\frac{1}{8}(x_1+x_3+x_4) +\frac{3}{8}=\frac{19}{64}\\
            &x_3'= -\frac{3}{8}x_1x_2x_4 -\frac{1}{8}(x_1x_2+x_1x_4+x_2x_4) +\frac{1}{8}(x_1+x_2+x_4) +\frac{3}{8}=\frac{33}{64}\\
            &x_4'= -\frac{3}{8}x_1x_2x_3 -\frac{1}{8}(x_1x_2+x_1x_3+x_2x_3) +\frac{1}{8}(x_1+x_2+x_3) +\frac{3}{8}=\frac{33}{64}
        \end{align*}

    \subsection{GradSAT}
        \texttt{GradSAT} construct the Boolean constraints with \ac{BDD}s~\cite{kyrillidis2021continuous}.
        By performing probabilistic inference on the \ac{BDD}, the constraint can be evaluated and differentiated.
        \begin{algorithm}[htbp]
            \caption{Differentiation with Belief Propagation}
            \textbf{Input}: Boolean formula $f$ with constraint set $C$ and the current assignment $x$\\
            \textbf{Parameter}: BDDs $B$ which encode the constraints \\
            \textbf{Output}: Gradient $\nabla_xF_f$
            \begin{algorithmic}[1]
                \State Initialize gradient $\nabla_xF_f=[0\;0\;\cdots\;0]$
                \State Initialize the top-down and bottom-up messages $M_{TD}$, $M_{BU}$ as 0. Except $M_{TD}(root)=M_{BU}(\texttt{true})=1$
                \For{$c\in{}C$}
                    \State Sort the nodes of $B_c$ literal by literal to a list $L$
                    \State For each node maintain an index $i$ to the literal
                    \For{each node $v\in{}L$}       \Comment{Forward traversal}
                        \State $M_{TD}[v.left]+=\frac{1-x[i_v]}{2}M_{TD}[v]$
                        \State $M_{TD}[v.right]+=\frac{1+x[i_v]}{2}M_{TD}[v]$
                    \EndFor
                    \For{each node $v\in{}L$}       \Comment{Backward traversal}
                        \For{each node $u$ such that $u.left=v$}
                            \State $M_{BU}[u]+=\frac{1-x[i_u]}{2}M_{BU}[v]$
                        \EndFor
                        \For{each node $u$ such that $u.right=v$}
                            \State $M_{BU}[u]+=\frac{1+x[i_u]}{2}M_{BU}[v]$
                        \EndFor
                        \State $\partial_{x[i_v]}F_f+=M_{TD}[v](M_{BU}[v.left]-M_{BU}[v.right])$
                    \EndFor
                \EndFor
                \State \Return $\nabla_xF_f$ 
            \end{algorithmic}
            \label{alg:gradsat}
        \end{algorithm}

        In the following, we give an example to demonstrate how the graph traversal in \texttt{GradSAT} works.

        \begin{example}
            (Evaluation by Forward Traversal)
            Given a constraint $x_1+x_2+x_3+x_4\ge{}2$, the \ac{BDD} can be constructed~\cite{een2006translating}.
            Given $x=(\frac{1}{2},\frac{1}{2},-\frac{1}{2},-\frac{1}{2})$, the forward traversal of \ac{BDD} can be described as:
            \begin{center}\begin{tikzpicture}[round/.style={circle, draw=black!100, thick, minimum size=8mm},
            trans/.style={rectangle, draw=black!0, thick, minimum width=1mm, minimum height=1mm},]
                \newcommand{\y}{1.}
                \newcommand{\x}{1.5}
                \node[round, label={above:$M_{TD}=1$},
                    ] (root) at (0, 0) {$x_1$};
                \node[round, label={above:$M_{TD}=\frac{1}{4}$}   ] (x2_1) at (2*\x  , \y) {$x_2^l$};
                \node[round, label={above:$M_{TD}=\frac{3}{4}$}   ] (x2_2) at (2*\x  ,-\y) {$x_2^r$};
                \node[trans, label={above:$M_{TD}=\frac{1}{16}$}  ] (x3_0) at (4.4*\x  , 2*\y) {\texttt{True}};
                \node[round, label={above:$M_{TD}=\frac{3}{8}$}   ] (x3_1) at (4*\x  , 0) {$x_3^l$};
                \node[round, label={above:$M_{TD}=\frac{9}{16}$}  ] (x3_2) at (4*\x  ,-2*\y) {$x_3^r$};
                \node[trans, label={above:$M_{TD}=\frac{9}{32}$}  ] (x4_0) at (6.4*\x  , \y) {\texttt{True}};
                \node[round, label={above:$M_{TD}=\frac{33}{64}$} ] (x4_1) at (6*\x  , -\y) {$x_4$};
                \node[trans, label={above:$M_{TD}=\frac{9}{64}$}  ] (x4_2) at (6.6*\x  ,-3*\y) {\texttt{False}};
                \node[trans, label={above:$M_{TD}=\frac{99}{256}$}] (true) at (8.4*\x  , 0) {\texttt{True}};
                \node[trans, label={above:$M_{TD}=\frac{33}{256}$}] (false) at (8.6*\x  ,-2*\y) {\texttt{False}};
                \newcommand{\arrowIn}{\tikz \draw[-stealth] (-1pt,0) -- (1pt,0);}
                \draw[thick] (root.north east) -- (x2_1.west) node[sloped,pos=0.5,allow upside down]{\arrowIn} node[midway, above]{$\frac{1}{4}$};
                \draw[thick, dashed] (root.south east) -- (x2_2.west) node[sloped,pos=0.5,allow upside down]{\arrowIn} node[midway, above]{$\frac{3}{4}$};
                \draw[thick] (x2_1.north east) -- (x3_0.west) node[sloped,pos=0.5,allow upside down]{\arrowIn} node[midway, above]{$\frac{1}{16}$};
                \draw[thick, dashed] (x2_1.south east) -- (x3_1.west) node[sloped,pos=0.5,allow upside down]{\arrowIn} node[midway, above]{$\frac{3}{16}$};
                \draw[thick] (x2_2.north east) -- (x3_1.west) node[sloped,pos=0.5,allow upside down]{\arrowIn} node[midway, above]{$\frac{3}{16}$};
                \draw[thick, dashed] (x2_2.south east) -- (x3_2.west) node[sloped,pos=0.5,allow upside down]{\arrowIn} node[midway, above]{$\frac{9}{16}$};
                \draw[thick] (x3_1.north east) -- (x4_0.west) node[sloped,pos=0.5,allow upside down]{\arrowIn} node[midway, above]{$\frac{9}{32}$};
                \draw[thick, dashed] (x3_1.south east) -- (x4_1.west) node[sloped,pos=0.5,allow upside down]{\arrowIn} node[midway, above]{$\frac{3}{32}$};
                \draw[thick] (x3_2.north east) -- (x4_1.west) node[sloped,pos=0.5,allow upside down]{\arrowIn} node[midway, above]{$\frac{27}{64}$};
                \draw[thick, dashed] (x3_2.south east) -- (x4_2.west) node[sloped,pos=0.5,allow upside down]{\arrowIn} node[midway, above]{$\frac{9}{64}$};
                \draw[thick] (x4_1.north east) -- (true.west) node[sloped,pos=0.5,allow upside down]{\arrowIn} node[midway, above]{$\frac{99}{256}$};
                \draw[thick, dashed] (x4_1.south east) -- (false.west) node[sloped,pos=0.5,allow upside down]{\arrowIn} node[midway, above]{$\frac{33}{256}$};
            \end{tikzpicture}\end{center}
            
            By summing up the top-down message ($M_{TD}$) at the leaf nodes, we can find the probability of this constraint being satisfied or unsatisfied by:
            \begin{align*}
                &P(true) = \sum{}M_{TD}(\texttt{true}) = \frac{1}{16} + \frac{9}{32}  + \frac{99}{256} =  \frac{187}{256}\\
                &P(false) = \sum{}M_{TD}(\texttt{false}) = \frac{33}{256} + \frac{9}{64}  =  \frac{69}{256}
            \end{align*}
            \label{ex:gradsat1}
        \end{example}
        
        Eg.~\ref{ex:gradsat1} explains the forward traversal process (Lines 6-8 of Alg.~\ref{alg:gradsat}), where the top-down messages $M_{TD}$'s are accumulated.
        Note that $x[i_v]\in[-1,1]$ encodes the truth value of a Boolean variable, as stated in Sec.~\ref{sec:pre_wfe}.
        In \texttt{GradSAT}, the probabilistic inference is propagating the probabilities, in which the randomized rounding is defined by $\mathbb{P}[x_i=\texttt{true}]=\frac{1-x_i}{2}$ and $\mathbb{P}[x_i=\texttt{false}]=\frac{1+x_i}{2}$ (Def. 4 in~\cite{kyrillidis2021continuous}).
        As a result, $M_{TD}$'s are propagating downstream with $M_{TD}(v.left)+=\mathbb{P}[v=\texttt{true}]M_{TD}(v)$ and $M_{TD}(v.right)+=\mathbb{P}[v=\texttt{false}]M_{TD}(v)$.
        And hence the top-down messages at leaf nodes ($M_{TD}(\texttt{true})$, $M_{TD}(\texttt{false})$) encode the probability of the constraint being satisfied or unsatisfied.
        
        \begin{example}
            (Differentiation by Backward Traversal)
            To compute the gradient, we need to perform backward traversal for the \ac{BDD}.
            \begin{center}\begin{tikzpicture}[round/.style={circle, draw=black!100, thick, minimum size=8mm},
            trans/.style={rectangle, draw=black!0, thick, minimum width=1mm, minimum height=1mm},]
                \newcommand{\y}{1.}
                \newcommand{\x}{1.5}
                \node[round, label={above:$M_{BU}=\frac{187}{256}$}] (root) at (0, 0) {$x_1$};
                \node[round, label={above:$M_{BU}=\frac{61}{64}$}   ] (x2_1) at (2*\x  , \y) {$x_2^l$};
                \node[round, label={above:$M_{BU}=\frac{21}{32}$}   ] (x2_2) at (2*\x  ,-\y) {$x_2^r$};
                \node[trans, label={above:$M_{BU}=1$}  ] (x3_0) at (4.4*\x  , 2*\y) {\texttt{True}};
                \node[round, label={above:$M_{BU}=\frac{15}{16}$}   ] (x3_1) at (4*\x  , 0) {$x_3^l$};
                \node[round, label={above:$M_{BU}=\frac{9}{16}$}  ] (x3_2) at (4*\x  ,-2*\y) {$x_3^r$};
                \node[trans, label={above:$M_{BU}=1$}  ] (x4_0) at (6.4*\x  , \y) {\texttt{True}};
                \node[round, label={above:$M_{BU}=\frac{3}{4}$} ] (x4_1) at (6*\x  , -\y) {$x_4$};
                \node[trans, label={above:$M_{BU}=0$}  ] (x4_2) at (6.6*\x  ,-3*\y) {\texttt{False}};
                \node[trans, label={above:$M_{BU}=1$}] (true) at (8.4*\x  , 0) {\texttt{True}};
                \node[trans, label={above:$M_{BU}=0$}] (false) at (8.6*\x  ,-2*\y) {\texttt{False}};
                \newcommand{\arrowIn}{\tikz \draw[-stealth] (-1pt,0) -- (1pt,0);}
                \draw[thick] (x2_1.west) -- (root.north east) node[sloped,pos=0.5,allow upside down]{\arrowIn} node[midway, above]{$\frac{61}{256}$};
                \draw[thick, dashed] (x2_2.west) -- (root.south east) node[sloped,pos=0.5,allow upside down]{\arrowIn} node[midway, above]{$\frac{63}{128}$};
                \draw[thick] (x3_0.west) -- (x2_1.north east) node[sloped,pos=0.5,allow upside down]{\arrowIn} node[midway, above]{$\frac{1}{4}$};
                \draw[thick, dashed] (x3_1.west) -- (x2_1.south east) node[sloped,pos=0.5,allow upside down]{\arrowIn} node[midway, above]{$\frac{45}{64}$};
                \draw[thick] (x3_1.west) -- (x2_2.north east) node[sloped,pos=0.5,allow upside down]{\arrowIn} node[midway, above]{$\frac{15}{64}$};
                \draw[thick, dashed] (x3_2.west) -- (x2_2.south east) node[sloped,pos=0.5,allow upside down]{\arrowIn} node[midway, above]{$\frac{27}{64}$};
                \draw[thick] (x4_0.west) -- (x3_1.north east) node[sloped,pos=0.5,allow upside down]{\arrowIn} node[midway, above]{$\frac{3}{4}$};
                \draw[thick, dashed] (x4_1.west) -- (x3_1.south east) node[sloped,pos=0.5,allow upside down]{\arrowIn} node[midway, above]{$\frac{3}{16}$};
                \draw[thick] (x4_1.west) -- (x3_2.north east) node[sloped,pos=0.5,allow upside down]{\arrowIn} node[midway, above]{$\frac{9}{16}$};
                \draw[thick, dashed] (x4_2.west) -- (x3_2.south east) node[sloped,pos=0.5,allow upside down]{\arrowIn} node[midway, above]{$0$};
                \draw[thick] (true.west) -- (x4_1.north east) node[sloped,pos=0.5,allow upside down]{\arrowIn} node[midway, above]{$\frac{3}{4}$};
                \draw[thick, dashed] (false.west) -- (x4_1.south east) node[sloped,pos=0.5,allow upside down]{\arrowIn} node[midway, above]{$0$};
            \end{tikzpicture}\end{center}
            
            With $M_{BU}(\texttt{true})$ as 1 and $M_{BU}(\texttt{false})$ as 0, the backward traversal is propagating the probability of the constraint being satisfied to the root node (Lines 9-13 of Alg.~\ref{alg:gradsat}, Thm.~6 of~\cite{kyrillidis2021continuous}): $M_{BU}(root) = P(\texttt{true})$. 
            If the bottom-up messages are initialized the other way around, i.e., with $M_{BU}(\texttt{true})$ as 0 and $M_{BU}(\texttt{false})$ as 1, $M_{BU}(root)$ should encode $P(\texttt{false})$.

            After performing both forward and backward traversals, both top-down and bottom-up messages ($M_{TD}$, $M_{BU}$) have been obtained.
            Then the partial derivatives can be computed by \footnote{Due to the difference in the definition of Boolean functions, the exact gradient computation method in~\cite{kyrillidis2021continuous} should modify Line 14 of Alg.~\ref{alg:gradsat} to $\partial_{x[i_v]}F_f+=\frac{1}{2}M_{TD}[v](M_{BU}[v.right]-M_{TD}[v.left])$. However, this does not bring about any essential difference since \texttt{FourierSAT} and \texttt{FastFourierSAT} is performing gradient descent, while \texttt{GradSAT} is performing gradient accent.}:
            \begin{align*}
                x_1'& = M_{TD}[x_1](M_{BU}[x_2^l]-M_{BU}[x_2^r]) = \frac{1}{2}\cdot{}1\cdot\left(\frac{21}{32}-\frac{61}{64}\right) = \frac{19}{64}\\
                x_2'& = M_{TD}[x_2^l](M_{BU}[true]-M_{BU}[x_3^l]) + M_{TD}[x_2^r](M_{BU}[x_3^l]-M_{BU}[x_3^r]) \\
                    &=\frac{1}{4}\cdot\left(1-\frac{15}{16}\right) + \frac{3}{4}\cdot\left(\frac{15}{16}-\frac{9}{16}\right) 
                    = \frac{1}{64} +\frac{9}{32} = \frac{19}{64}\\
                x_3'& = \frac{1}{2}M_{TD}[x_3^l](M_{BU}[true]-M_{BU}[x_4]) + M_{TD}[false](M_{BU}[x_3^r]-M_{BU}[x_4])\\ 
                    &= \frac{1}{4}\cdot\left(1-\frac{3}{4}\right) + \frac{9}{16}\cdot\left(\frac{3}{4}-0\right) 
                    = \frac{1}{16} + \frac{27}{64} = \frac{33}{64}\\
                x_4'& = M_{TD}[x_4](M_{BU}[true]-M_{BU}[false]) =  \frac{1}{2}\cdot{}\frac{33}{64}\cdot\left(1-0\right) = \frac{33}{64}\\
            \end{align*}
            Eventually, the resulting gradient is $x' = \left(\frac{19}{64},\frac{19}{64},\frac{33}{64},\frac{33}{64}\right)$, which matches \texttt{FourierSAT} (Eg.~\ref{ex:fouriersat}) and \texttt{FastFourierSAT} (Eg.~\ref{ex:backward}).
            \label{ex:gradsat2}
        \end{example}

    \subsection{Projected Gradient Descent}\label{app:PGD}
    At each iteration in \ac{CLS}, the optimizer tries to find a local minimum of the weighted objective function (Eq.~\ref{eq:objective}).
    In gradient descent, using the first order method to update the variable $x_i$'s can be described as:
    \begin{equation*}
        x_{k+1} = x_k - \eta\sum_{c\in{}C}w_c\nabla{FE}_c(x_k)
    \end{equation*}
    where $\eta$ is the step size of gradient descent and $\nabla(\cdot)$ is the is a differential operator.

    In \ac{CLS}-based SAT solving, the variables are bounding within $[-1,1]^n$ due to Thm.~\ref{thm:walshexpansion}.
    And hence the optimization problem becomes:
    \begin{equation}
        x^* = \mathop{\arg\min}\sum_{c\in{}C_f}w_cFE_c(x) + I_{[-1,1]^n}(x)
        \label{eq:PGFP}
    \end{equation}
    where $I_{[-1,1]^n(x)}$ is the indication function, which is equal to 0 when $x\in[-1,1]^n$, otherwise equal to $+\infty$.
    Since the indication function is not differentiable, a \ac{PGFP} algorithm is utilized to solve Eq.~\ref{eq:PGFP} implicitly~\cite{griewank2008evaluating,krantz2002implicit}.
    \ac{PGFP} iteratively updates the solution and applies the proximal operator, which we denote as $z$ and $\Pi$, respectively.
    Using the Moreau-Yosida regularization~\cite{parikh2014proximal, niculae2017regularized}, the proximity operator is given as:
    \begin{equation*}
        \Pi(z)= \mathop{\arg\min}\limits_{x\in[-1,1]^n}
        \lVert{x-z}\rVert_2^2
    \end{equation*}
    As we related the proximal gradient with gradient descent as $z=x_{k+1}$, the optimization problem in \ac{CLS}-based SAT can be derived as:
    \begin{equation*}
        x_{k+1} = \mathop{\arg\min}\limits_{x\in[-1,1]^n}
        \lVert{x - x_k - \eta\sum_{c\in{}C_f}\nabla{FE}_c(x_k)}\rVert_2^2
    \end{equation*}
    And hence the gradient descent in \ac{CLS} can be described as Alg.~\ref{alg:pgfp}.
    
    \begin{algorithm}[htbp]
        \caption{Using projected gradient descent to update the variables.}
        \label{alg:pgfp}
        \textbf{Input}: The current assignment $x$, the evaluation function $F(\cdot)$ and gradient function $G(\cdot)$\\
        \textbf{Parameter}: The current step size for gradient descent $\eta$\\
        \textbf{Output}: The updated assignment $x$
        \begin{algorithmic}[1]
            \For{$j = 1, \cdots, T$}
                \State $x, \eta=\text{lineSearch}\left(F(\cdot),G(\cdot),x,\eta\right)$      \Comment{Default method in~\cite{blondel2022efficient} is FISTA~\cite{beck2009fast}}
                \State $x=\text{clip}(x,-1,1)$   \Comment{\ac{PGFP} in \ac{CLS}}
                \If{$\eta<10^{-12}$}
                    \State \Return $x$
                \EndIf
            \EndFor
            \State \Return $x$
        \end{algorithmic}
   \end{algorithm}

   \texttt{FourierSAT}, \texttt{GradSAT}, and \texttt{FastFourierSAT} apply different evaluation $F(\cdot)$ and differentiation $G(\cdot)$ methods to Alg.~\ref{alg:pgfp}. 
   The line search in Alg.~\ref{alg:pgfp} needs to iteratively call $F(\cdot)$ and $G(\cdot)$ until either the step size $\eta$ is smaller than a threshold or the maximum iteration limit is reached, whichever occurs first.
    
\section{Transform-Free Due to Zero Entries}\label{app:xor}
    The Walsh coefficient of some constraints might have zero entries, making part of the computation trivial.
    By eliminating the trivial operations, the complexity can be further reduced.
    For example, the Walsh coefficient of an XOR constraint is $\hat{f}_{XOR}=[1\;0\;\cdots\;0]$, \emph{i.e.,} the Walsh expansion only has the highest order term and all other entries are 0.
    Then, the conjugated Walsh coefficient becomes $\tilde{f}_{XOR}=\hat{f}_{XOR}\cdot{}W^{-1}=[1\;1\;\cdots\;1]$.
    By writing out and rearranging Eq.~\ref{eq:inverse}, the Walsh expansions of XOR constraints can be evaluated as:
    \begin{equation}
        {WE}_{XOR}=\frac{1}{k+1}\Big({\sum_{a=0}^k\omega^{ka} + \sum_{a=0}^{k}\omega^{(k-1)a}\sum_{b=0}^kx_b + \cdots + \sum_{a=0}^{k}\omega^{0a}\prod_{b=0}^kx_b}\Big) 
        \label{eq:xor}
    \end{equation}
    It can be proved by the Euler formula and trigonometric identities that, $\sum_{a=0}^{k}\omega^{ab}=0,\;\forall b\in\mathbb{Z}^+$.
    By eliminating the trivial computations, \texttt{FastFourierSAT} will equivalently bypass the transform operators 
    \footnote{\cite{ercsey2011optimization} takes $s_i=(1+x_i)/2$. The computation of Walsh expansion of a CNF clause can also be reduced to multiplications} (\emph{step i and iii}).
    Then, the complexity only needs to account for traversing the binary tree forward and backward (\emph{step ii}).
    \begin{corollary}
        (Reduction) For XOR constraints, the complexity of running Autodiff for Alg.~\ref{alg:forward} can be reduced to $O(k)$.
        \label{cor:reduction}
    \end{corollary}

\section{Restart Heuristics}\label{app:heuristic}
\begin{figure}[htbp]
        \centerline{\includegraphics[width=0.6\textwidth]{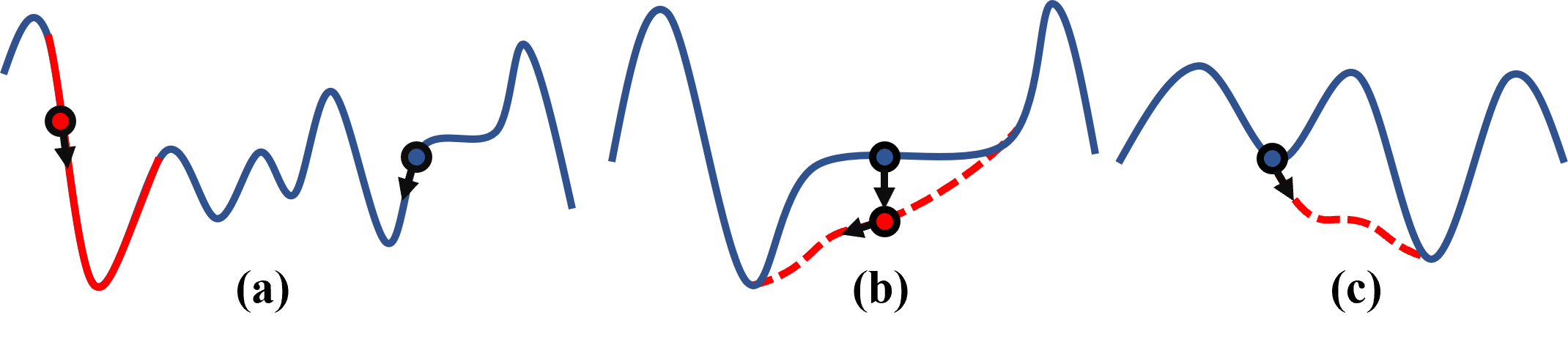}}
        \caption{(a) Different initialization points will converge to different local optima.
                 When the search trajectory is stuck at (b) a saddle point or (c) a local optimum, the weighting heuristic adapts the weights.}
        \label{fig:cls}
    \end{figure}

    The essence of \ac{CLS}-based hybrid SAT solving is using convex optimization to find the ground states of the non-convex energy landscape.
    Hence, the optimization results depend heavily on the initialization~\cite{jain2013low}.
    A \ac{CLS} approach can find a global optimum only if the neighboring convex set of the initialization point (\emph{e.g.}, the red dot in Fig.~\ref{fig:cls}a) includes a solution to the hybrid SAT formula, such that the local optimum searched by gradient descent will be the global optimum.
        
    Hence, the solution quality of CLS can be significantly improved by using random restart. 
    By leveraging the parallelism of GPUs, one can instantiate a batch of kernels to optimize the objective function with a batch of random initialization points. 
    Instead of blind random restarts, we propose to incorporate weighting and rephasing heuristics into the parallel search to help \texttt{FastFourierSAT} explore the search space in a more efficient way.
    
    For the weighting heuristic, we choose $\alpha{}=0.4$ by convention~\cite{liang2016exponential}.
    From the below examples, we can observe how the adaptive weight can escape the saddle points or local optima.
    \begin{example}
        (Saddle points)
        Consider a simple example that only has two clauses $\{c_1=x_1\vee\neg{}x_2,\, c_2 = \neg{}x_1\oplus{}x_2\}$, where the corresponding objective function is $F=-\frac{1}{2}+\frac{x_1}{2}-\frac{x_2}{2}-\frac{3x_1x_2}{2}$. 
        At $\left(-\frac{1}{3},\,\frac{1}{3}\right)$, the optimization stops at an inner saddle point (blue dot in Fig.~\ref{fig:cls}b with unsatisfaction scores of $r_1 = 0$ and $r_2 = 1$.
        The weights will then adapt to $w'_1 = 0.6$, $w'_2 = 1$, resulting in a new objective function $F'=-\frac{3}{10}+\frac{3x_1}{10}-\frac{3x_2}{10}-\frac{13x_1x_2}{10}$.
        As a result, the gradient becomes $(\partial_{x_1}F', \partial_{x_2}F') = (-\frac{3}{4}, \frac{3}{4})$ (red dot in Fig.~\ref{fig:cls}b).
        \label{ex:saddle}
    \end{example}
    \begin{example}
        (Local optima)
        Consider another simple example $\{c_1=x_1\oplus{}x_2,\, c_2 = x_2\oplus{}x_3,\, c_3 = x_3\oplus{}x_4\}$, where the corresponding objective function is $F=x_1x_2+x_2x_3+x_3x_4$. 
        At $\left(1, \,-1, \,-1, \,1\right)$, the optimizer stops at the boundary (blue dot at Fig.~\ref{fig:cls}c) with unsatisfaction scores of $r_1=0$, $r_2=1$, $r_3=0$.
        The weights will then adapt to $w'_1 = w'_3= \frac{3}{5}$, $w'_2 = 1$, resulting in a new objective function $F'=\frac{3}{5}x_1x_2+x_2x_3+\frac{3}{5}x_3x_4$.
        As a result, the gradient becomes $(\partial_{x_1}F', \partial_{x_2}F', \partial_{x_3}F', \partial_{x_4}F') = (-\frac{3}{5}, -\frac{2}{5}, -\frac{2}{5}, -\frac{3}{5})$ and the search trajectory will move away from the boundary at $x_2=x_4=-1$.
        \label{ex:local}
    \end{example}

    Usually, the weighting heuristic is insufficient.
    After escaping from the saddle point (local optimum), the \ac{CLS} can fall into another saddle point (local optimum).
    In Eg.~\ref{ex:saddle}, the adaptive weights only move the saddle point closer to a local optimum, and \ac{CLS} still searches along that direction. 
    In Eg.~\ref{ex:local}, \ac{CLS} may bounce between two local optima when $x_2$ is always equal to $x_3$.
    Hence, we propose to utilize the rephasing heuristic jointly with the weighting heuristic to mitigate the aforementioned issue.
    For the rephasing heuristic, we want to balance between ``intensification'' and ``diversification''.
    Hence, we propose to adopt a policy of $(ROF)^\infty$. 
    
    To study how the heuristic can effectively improve CLS, we choose 500 random 3-SAT problems from SATLIB~\cite{hoos2000satlib}, with $n \in \{50, 100, 150, 200, 250\}$.
    These ``toy'' problems are easy for the SOTA SAT solvers. However, due to the high clause-to-variable ratio, the solution densities of these problems are low. 
    These problems are also suitable for studying the search efficiency of \ac{CLS} because the number of global minima is much smaller than the local minima. 
    We compare the performance of \texttt{FastFourierSAT} with and without heuristic, and the portfolio of the two. 
    We set the thread-level parallelism to 1024 and observe the number of instances that can be solved within 1000 iterations.
    The results are shown in Table~\ref{tab:ablation}, where it takes \verb|FastFourierSAT| more iterations to solve larger problems.
    With the dedicated heuristics, \texttt{FastFourierSAT} can achieve 3.38$\times$ speedup for $n=150$ and can solve 21 more instances in total.
    The most effective strategy may be varied for different problems.
    For example, when the optimal solutions are sparse and distributed in the search space, it is preferable to diversify the search instead of intensify.
    Employing multiple GPUs to simultaneously attempt different search strategies can, hence, achieve better performance. 
    Thus, the portfolio of two achieves a further 1.15$\times$ speedup and can solve 7 more instances.

    \begin{table*}[tb]
        \addtolength{\tabcolsep}{-3pt}
        \centering
        \begin{tabular}{ccccccccccccccccccccc}
        \hline
        &  & \multicolumn{9}{c}{Average Iterations (Par 2)} &  & \multicolumn{9}{c}{Number of Instances Solved} \\ \cline{3-11} \cline{13-21} 
        \multirow{-2}{*}{Methods} &  & 50 &  & 100 &  & 150 &  & 200 &  & 250 &  & 50 &  & 100 &  & 150 &  & 200 &  & 250 \\
        \cline{1-1} \cline{3-3} \cline{5-5} \cline{7-7} \cline{9-9} \cline{11-11} \cline{13-13} \cline{15-15} \cline{17-17} \cline{19-19} \cline{21-21} 
        w/o   heuristics &  & 1.02 &  & 2.01 &  & 74.03 &  & 533.6 &  & 1075 &  & 100 &  & 100 &  & 98 &  & 81 &  & 53 \\
        w/ heuristics &  & 1.01 &  & 2.2 &  & 21.03 &  & 405.8 &  & 797.8 &  & 100 &  & 100 &  & 100 &  & 85 &  & 68 \\
        Portfolio &  & 1.01 &  & 1.58 &  & 15.9 &  & 322.4 &  & 724.1 &  & 100 &  & 100 &  & 100 &  & 89 &  & 71 \\
        \hline
        \end{tabular}
        \caption{Results on solving weighted 3-SAT problems from SATLIB.}
        \label{tab:ablation}
    \end{table*}

\section{Benchmark Details}\label{app:benchmark}
    \paragraph{Random Boolean Formula.}
        Seven random boolean formulas are generated for studying the gradient computation speeds in \emph{RQ1.}
        \begin{itemize}
            \item \texttt{xor+card}: 800 XOR constraints, each has a length of 8. And a cardinality constraint with a length of 32.
            \item \texttt{xor1}: 200 XOR constraints, each has a length of 8.
            \item \texttt{xor2}: 400 XOR constraints, each has a length of 16.
            \item \texttt{xor3}: 800 XOR constraints, each has a length of 32.
            \item \texttt{card1}: 50 cardinality constraints, each has a length of 8.
            \item \texttt{card2}: 100 cardinality constraints, each has a length of 16.
            \item \texttt{card3}: 200 cardinality constraints, each has a length of 32.
        \end{itemize}
    We evaluate the above Boolean constraints with 10000 random points and compare among \texttt{FourierSAT}, \texttt{GradSAT}, and the proposed \texttt{FastFourierSAT}.
    The results in Fig.~\ref{fig:grad} show the average gradient computation time per random point.

    In the following, we describe the benchmarks and competing solvers used to study the performance of \texttt{FastFourierSAT} in \emph{RQ2} and \emph{RQ3}.  
    
    \paragraph{Benchmark 1: Random cardinality formula.}
        We chose $N\in\{50, 100, 150, 200, 250\}$, $l=0.2N$ and $m=0.6N$ to generate $m$ random $\frac{l}{2}$-cardinality constraints.
        Each constraint randomly sampled $l$ variables as the literals with 0.5 probability of all positive or all negative.
        For each $N$, 100 instances were generated.
                
    \paragraph{Benchmark 2: Parity learning with error.}
        Parity learning with error aims to identify an unknown Boolean function that can satisfy at least $(1-e)m$ I/O samples when $m$ I/O samples are given.
        For $0<e<\frac{1}{2}$, whether the problem is in P still remains an open question, and is known to be hard for local search solvers~\cite{crawford1994minimal}.
        We chose $N \in \{20, 30, 40, 50, 60\}$, $e = \frac{1}{4}$ and $m = 2N$ to generate hard instances.
        For the \ac{CLS} solvers, this problem can be encoded into solving $m$ XOR clauses with at most $em$ clauses that can be violated.
        For the other solvers, we used the encoding due to~\cite{hoos2000satlib}.
    
    \paragraph{Benchmark 3: Weighted Max-Cut problem.}
        The weighted max-cut problem is a combinatorial optimization problem that involves partitioning the nodes of a weighted graph into two disjoint sets in such a way that the sum of the weights of the edges crossing the partition is maximized. 
        This problem is known to be NP-hard.
        We collect the weighted Max-Cut instances (total 102) from MSE 2019.
        Additionally, We have generated random planted partition graphs with a specified number of clusters ($l$) and vertices per cluster ($k$).
        Specifically, we choose $(l,k) = \{(16,16),(16,32),(32,16),(32,32)\}$ to generate the weighted Max-Cut problems.
        Within each cluster, the edges were assigned a weight of 1, while edges between different groups had a weight of 2.
        For each $(l,k)$, we generated 100 instances.

        Benchmark 3 is an optimization problem, we evaluate the performance of the solvers based on the relative score, where for solvers s on i-th instance is calculated using:
        \begin{equation}
            \nonumber
            score(s,i) = \frac{\max_s(cost(s, i))-cost(s, i)+1}{\max_s(cost(s, i))-\min_s(cost(s, i))+1}
        \end{equation}
        When the solvers are top-performing, the scores should close to 100\%, and 0\% otherwise.

    In the above benchmark, only \texttt{GradSAT} and \texttt{FastFourierSAT} can natively accept XOR and cardinality constraints.
    For other solvers, we encoded the Boolean formulas into CNF or weighted CNF (WCNF).
    Specifically, the cardinality constraints are encoded by~\cite{ansotegui2021optilog}, which automatically chooses the best encoding scheme.
    The XOR constraints are encoded by~\cite{li2000integrating}.
    First, a $k$-XOR constraint is decomposed into $3$-XOR constraints, which are then subsequently encoded into $3$-CNF format.
    The Benchmark~3 only consists of $2$-XOR constraints which can be directly encoded into $2$-CNF format.
    
    \paragraph{Competing Solvers}
        The evaluation of the above benchmarks is given a time limit of 60 seconds.
        All experiments were conducted on high-performance computer cluster nodes.
        The CPU compute node is configured with dual AMD EPYC 7773X CPUs.
        The GPU compute node is configured with dual AMD EPYC 7713 CPUs and NVIDIA A100 SXM4 40GB GPUs. 
        Each solver was executed on 32 CPU threads with 2 GB of RAM per thread (and 1 GPU if necessary).

        We compare the CPU and GPU implementation of \ac{CLS} using \texttt{GradSAT} and \texttt{FastFourierSAT}.
        For Benchmarks 1 and 2, the following parallel SAT solvers are included:
        \begin{itemize}
            \item \texttt{PalSAT}~\cite{Biere-SAT-Competition-2017-solvers}: the parallel version of \verb|YalSAT| (a \ac{DLS} solver) which is the best solver of SAT Competition 2017 in the ``Random Track'' category. 
            \item \texttt{P-MCOMSPS-STR-SC}~\cite{li2021cadical}: the best solver of SAT Competition 2021 in the ``Main Parallel Track'' category.
            It uses \texttt{Painless} framework~\cite{le2017painless} to parallelize many single-threaded solvers, in which the learned clause can be shared with each other.
            The base single-threaded solver is a \ac{CDCL} solver \texttt{MapleCOMSPS}~\cite{liang2017maple}.
            We refer this to \texttt{Painless+Maple} in the main text.
            \item \texttt{ParKissat-RS}~\cite{zhang2022parkissat}: the best solver of SAT Competition 2022 in the ``Main Parallel Track'' category.
            It is similar to the previous solver but uses a hybrid solver \texttt{Kissat-MAB}~\cite{cherif2021kissat} as the base single-threaded solver.
            We refer this to \texttt{Painless+Kissat} in the main text.
            \item \texttt{GpuShareSat}~\cite{prevot2021leveraging}: a GPU augmented SAT solving approach which enables the clause sharing between base single-threaded solvers.
            It uses \ac{CDCL} solver \texttt{GlucoseSyrup}~\cite{audemard2014lazy} or hybrid solver \texttt{RelaxedLCMDCBDLnewTech}~\cite{cai2021deep} as the base single-threaded solvers.
            We refer these to \texttt{GSS+Glucose} and \texttt{GSS+RelaxedSAT} in the main text.
            \item \texttt{ParaFrost}~\cite{osama2021sat}: a GPU augmented SAT solving approach which applies modern inprocessing techniques in parallel.
            Note that it uses only one enhanced solver, whereas all other solvers in this benchmark are using a portfolio of many base solvers.
            \item \texttt{VBS w/o CLS}: The virtual best solver of the listed solvers above.
        \end{itemize}

        For Benchmark 3, we compare our method with the following Max-SAT solvers entered recent \ac{MSE}:
        \begin{itemize}
            \item \texttt{TT-Open-WBO}~\cite{nadel2020tt}: the best solver in MSE 2020 in the ``Weighted Incomplete Track 60 s'' category.
            It is also the base \ac{CDCL} solver for \verb|NuWLS-c| and \verb|SATLike-ck|.
            \item \texttt{SATLike-ck}~\cite{lei2021satlike}: the best solver in MSE 2021 in the ``Weighted Incomplete Track 60 s'' category.
            \item \texttt{SATLike}~\cite{cai2020old}: the base \ac{DLS} solver used in \verb|SATLike-ck|.
            \item \texttt{NuWLS-c}~\cite{chu2022nuwls}: the best solver in MSE 2022 in the ``Weighted Incomplete Track 60 s'' category. 
            \item \texttt{NuWLS}~\cite{chu2023nuwls}: the base \ac{DLS} solver used in \verb|NuWLS-c|.
            \item \texttt{VBS w/o CLS}: The virtual best solver of the listed solvers above.
        \end{itemize}

\section{Implementation Details}\label{app:imple}

    The hyperparameters used in the benchmark are listed below:
    \begin{itemize}
        \item $\alpha$: the decay factor used in Prop.~\ref{prop:erwa}. We choose $\alpha=0.4$ by convention~\cite{liang2016exponential}.
        \item $p_t$: the thread-level parallelism described in Sec.~\ref{sec:gpu}. It is the number of parallel tasks launched with different initial assignments concurrently. Different benchmarks choose a different $p_t$ to maximize leveraging the parallelism of the GPU.
        \item $\Pi$: the rephasing policy in Sec.~\ref{sec:rephase}. At each restart, the policy chooses a phase from \emph{i. \underline{O}riginal phase}, \emph{ii, \underline{F}lipped phase}, and \emph{iii, \underline{R}andom phase}. Different benchmarks choose different policies.
    \end{itemize}
    
    In Benchmark~1, we did not choose high thread-level parallelism since the computation requires double precision (otherwise incurs a numerical round-off error with the transform).
    A thread-level parallelism $p_t=32$ is chosen.
    We use the adaptive weighting heuristic as described in the main text with a policy of $(ROF)^\infty$ for rephasing heuristics.

    The parity learning with error problem in Benchmark~2 can be formulated as optimizing XOR constraints. 
    The complexity of computing the gradient of the XOR constraints is $O(k)$ with $k$ literals (see Appx.~\ref{app:xor}), while the complexity for the cardinality constraint in benchmark 1 is $O(k^2)$.
    And hence we choose thread-level parallelism $p_t=1024$ ($=32^2$) for this benchmark.
    For optimization problems, the Thm.~\ref{thm:certificate} does not hold. 
    When applying adaptive weight, the local optima which encode the solutions might have higher energy than the non-solution local optima.
    Therefore, we use fixed weights during the local search.
    Consequently, since the fixed weights do not deform the energy landscape and there is no reason to intensify searching the local optima, the rephasing heuristic is modified to $(RF)^\infty$.
    We use a policy of $(RF)^\infty$ for rephasing heuristics instead.
        
    Similarly, the weighted Max-Cut is also an optimization problem.
    We use a similar configuration as Benchmark~2 for the restart heuristics but we chose a different thread-level parallelism with $p_t=32$ because the number of variables and clauses in this benchmark are much larger than the previous benchmarks.

\section{Technical Proofs}
    \subsection{Recap of Theorem~\ref{thm:conv}}
        Let's denote two one-dimension sequences as $g_i=[x_{i0}, x_{i1},\cdots,x_{in}]^T$ and $g_j=[x_{j0}, x_{j1},\cdots,x_{jm}]^T$, where $n<m$.
        To prove Eq.~\ref{eq:convolution} is to show the equivalence between $\mathcal{F}\left(g_i*g_j\right)$ and $\mathcal{F}\left(g_i\right)\mathcal{F}\left(g_j\right)$.
        We first show $\mathcal{F}\left(g_i*g_j\right)$:
        \begin{equation*}
            g_i*g_j=\left[
                x_{i0}x_{j0},  \cdots,  \sum_{q=0}^{k}x_{iq}x_{j(k-q)},  \cdots,  x_{in}x_{jm}
            \right]^T
        \end{equation*}
        Then the convoluted sequence in the Fourier domain will be:
        \begin{align}
            \mathcal{F}(g_i*g_j)=\left[
                \sum_{k=0}^{n+m}\sum_{q=0}^{k}x_{iq}x_{j(k-q)},  \cdots,  \sum_{k=0}^{n+m}\omega^{pk}\sum_{q=0}^{k}x_{iq}x_{j(k-q)},  \cdots,  \sum_{k=0}^{n+m}\omega^{{(n+m)k}}\sum_{q=0}^{k}x_{iq}x_{j(k-q)}
            \right]^T
            \label{eq:proof_convolved}
        \end{align}
        Then we show $\mathcal{F}\left(g_i\right)\mathcal{F}\left(g_j\right)$:
        \begin{align*}
            \mathcal{F}(g_i) &= \left[
                \sum_{q=0}^nx_{iq},  \cdots,  \sum_{q=0}^n\omega^{pq}x_{iq},  \cdots,  \sum_{q=0}^n\omega^{(n+m)q}x_{iq}
            \right]^T\\
            \mathcal{F}(g_j) &= \left[
                \sum_{q=0}^mx_{jq},  \cdots,  \sum_{q=0}^m\omega^{pq}x_{jq},  \cdots,  \sum_{q=0}^m\omega^{(n+m)q}x_{jq}
            \right]^T
        \end{align*}
        Then the multiplied sequence in the Fourier domain will be:
        \begin{align}
            &\mathcal{F}(g_i)\circ\mathcal{F}(g_j)\label{eq:proof_multiplied}\\
            =&\left[
                \left(\sum_{q=0}^nx_{iq}\right)\left(\sum_{q'=0}^mx_{jq'}\right),  \cdots,  \left(\sum_{q=0}^n\omega^{pq}x_{iq}\right)\left(\sum_{q'=0}^m\omega^{pq'}x_{jq'}\right),  \cdots,  \left(\sum_{q=0}^n\omega^{(n+m)q}x_{iq}\right)\left(\sum_{q'=0}^m\omega^{(n+m)q'}x_{jq'}\right)
            \right]^T\nonumber
        \end{align}
        The $p$-th term in Eq.~\ref{eq:proof_multiplied} can be written as:
        \begin{equation*}
            \left(\sum_{q=0}^n\omega^{qp}x_{iq}\right)\left(\sum_{q'=0}^m\omega^{q'p}x_{jq'}\right) = \sum_{q=0}^n\sum_{q'=0}^m\omega^{(q+q')p}x_{iq}x_{jq'}\\
        \end{equation*}
        Consider $k=q+q'$:
        \begin{equation*}
            \sum_{q=0}^n\sum_{q'=0}^m\omega^{(q+q')p}x_{iq}x_{jq'}= \sum_{k=0}^{n+m}\omega^{pk}\sum_{q=0}^{k}x_{iq}x_{j(k-q)}
        \end{equation*}
        As a result, the $p$-th term in Eq.~\ref{eq:proof_multiplied} is equivalent to $p$-th term in Eq.~\ref{eq:proof_convolved}.
        This result can be generalized to any $p$, and hence $\mathcal{F}\left(g_i*g_j\right)$ is equivalent to $\mathcal{F}\left(g_i\right)\circ\mathcal{F}\left(g_j\right)$.

        Besides, the proof above also shows:
        \begin{corollary}
            (Communitivity) 
            With $g$ and $h$ as two one-dimension sequences.
            The linear convolutions ($*$) in space and the pointwise multiplications ($\circ$) in the Fourier domain are commutative.
            \begin{equation*}
                \mathcal{F}(g*h) \Leftrightarrow \mathcal{F}(h*g) \Leftrightarrow 
                \mathcal{F}(g)\circ\mathcal{F}(h) \Leftrightarrow 
                \mathcal{F}(h)\circ\mathcal{F}(g)
            \end{equation*}
            \label{cor:commute}
        \end{corollary}
        
    \subsection{Proof of Corollary~\ref{cor:asso}}
        Let's denote the one-dimension sequences in space as $g_i=[x_{i0}, x_{i1},\cdots,x_{in}]^T$.
        To prove Eq.~\ref{eq:asso} is to show the equivalence between $\mathcal{F}\left(g_1*\cdots{}*g_k\right)$ and $\mathcal{F}\left(g_1\right)\circ\cdots\circ\mathcal{F}\left(g_k\right)$.
        
        Due to Corollary~\ref{cor:commute}, the second term can be written as:
        \begin{align}
            \nonumber
            &\mathcal{F}\left(g_1\right)\circ\mathcal{F}\left(g_2\right)\circ\mathcal{F}\left(g_3\right)\circ\cdots\circ\mathcal{F}\left(g_k\right)\\
            \nonumber
            =&\left(\left(\left(
                        \mathcal{F}\left(g_1\right)\circ\mathcal{F}\left(g_2\right)
                    \right) \circ\mathcal{F}\left(g_3\right)
                \right)\circ\cdots
            \right)\circ\mathcal{F}\left(g_k\right)\\
            \nonumber
            =&\left(\left(
                    \mathcal{F}\left(g_1*g_2\right)\circ\mathcal{F}\left(g_3\right)
                \right)\circ\cdots
            \right)\circ\mathcal{F}\left(g_k\right)\\
            =&\left(
                    \mathcal{F}\left(g_1*g_2*g_3\right)\circ\cdots
            \right)\circ\mathcal{F}\left(g_k\right)
            \label{eq:proof_associative}
        \end{align}
        Eventually, Eq.~\ref{eq:proof_associative} will be equivalent to the first term.

    \subsection{Proof of Theorem~\ref{thm:certificate}}
        Definition~\ref{def:objective} defines the objective function of a Boolean formula $F_f$ with a constraint set $C$ represented with Walsh expansions.
        Due to Theorem~\ref{thm:walshexpansion}, Eq.~\ref{eq:objective} can take value from $-\sum_{c\in{}C}w_c$ to $\sum_{c\in{}C}w_c$.
        \begin{itemize}
            \item ``$\Rightarrow$'': 
            Suppose the Boolean formula is satisfiable and an assignment $a^*\in\{\texttt{True}, \texttt{False}\}^n$ satisfies all the constraints.
            Correspondingly, we will have a ground state $x^*\in[-1,1]^n$ encodes $a^*$.
            For $c\in{}C$, $WE_c(x^*)=-1$ due to Theorem~\ref{thm:walshexpansion} and hence, $F_f(x^*)=-\sum_{c\in{}C}w_c$.
            Therefore, the minimum value of Eq.~\ref{eq:objective} is obtained.
            \item ``$\Leftarrow$'': 
            Suppose the minimum value of Eq.~\ref{eq:objective} is attainable.
            Thus, $\exists{x^*}$ such that $F_f(x^*)=-\sum_{c\in{C}}w_c$.
            Since $WE_c(x^*)$ can only take value from $-1$ to $1$, Eq.~\ref{eq:objective} can be minimal only if $WE_c(x^*)=-1, \forall{}c\in{}C$.
            Note that $WE_c(x^*)=-1$ encodes \texttt{True} of a constraint $c$, and hence all constraints in the constraint set $C$ is satisfied.
        \end{itemize}

    \subsection{Proof of Theorem~\ref{thm:forward}}\label{app:forward}
        In section~\ref{sec:forward} we have shown that the three-step process can evaluate the Walsh expansion of a symmetric Boolean constraint.
        Given a variable $x_i\in\{x_1, x_2, \cdots, x_k\}$, the \ac{ESPs} can be computed as Eq.~\ref{eq:convolution}.
        
        Step~i is performing the Fourier transform on the sequences in \ac{RHS} of Eq.~\ref{eq:convolution}.
        The overall computation can be simplified as the outer additions of a column vector and a row vector as in Eq.~\ref{eq:forward}.
        Since the column vector has a length of $k+1$ and the row vector has a length of $k$, $k(k+1)$ additions are required.
        Hence the complexity of this step is in $O(k^2)$.
        
        Step~ii is performing reduce product along the row, in which the length is $k$ and requires $k$ multiplications for each.
        So there will be $k(k+1)$ multiplications for all the rows in this step.
        
        In the end, step~iii uses Eq.~\ref{eq:inverse} to evaluate the Walsh expansion.
        The vector-vector multiplication consists of $k+1$ multiplications and $k$ additions.
        
        We can observe that the floating point operations of the overall procedure consist of $(k+1)^2$ additions and $(k+1)^2$ multiplication.
        Therefore, the complexity of this algorithm scales at $O(k^2)$.

    \subsection{Proof of Theorem~\ref{thm:backward}}\label{app:backward}
        Example~\ref{ex:forward} and Example~\ref{ex:backward} showcase the gradient computation process in \texttt{FourierSAT}.
        The computation graph is constructed based on Alg.~\ref{alg:forward}.
        The Walsh expansion $WE_c$ and the gradients $x$ can be obtained by forward and backward traversal of the graph.
        In Appendix~\ref{app:forward} we have seen the complexity of forward traversal is $O(k^2)$.

        For analyzing the backward traversal, we first write out all the terms Eq.~\ref{eq:autodiff} as:
        \begin{align}
            \nonumber&\gamma_c'=\frac{\partial{}WE_c}{\partial{}\gamma_c}=\tilde{f}_c\\
            \nonumber&\gamma_{1:2}'=\gamma_c'\circ\frac{\partial{}\gamma_c}{\partial{}\gamma_{1:2}}=\tilde{f}_c\circ\gamma_{3:4}\\
            \nonumber&\gamma_{3:4}'=\gamma_c'\circ\frac{\partial{}\gamma_c}{\partial{}\gamma_{3:4}}=\tilde{f}_c\circ\gamma_{1:2}\\
            \nonumber&\gamma_1'=\gamma_{1:2}'\circ\frac{\partial{}\gamma_{1:2}}{\partial{}\gamma_2}=\gamma_{1:2}'\circ\gamma_2\\
            \nonumber&\gamma_2'=\gamma_{1:2}'\circ\frac{\partial{}\gamma_{1:2}}{\partial{}\gamma_1}=\gamma_{1:2}'\circ\gamma_1\\
            \label{eq:chain}&\gamma_3'=\gamma_{3:4}'\circ\frac{\partial{}\gamma_{3:4}}{\partial{}\gamma_4}=\gamma_{3:4}'\circ\gamma_4\\
            \nonumber&\gamma_4'=\gamma_{3:4}'\circ\frac{\partial{}\gamma_{3:4}}{\partial{}\gamma_3}=\gamma_{3:4}'\circ\gamma_3\\
            \nonumber&x_1'=\frac{\partial{}\gamma_1}{\partial{}x_1}=\mathds{1}\cdot\gamma_i'\\
            \nonumber&x_2'=\frac{\partial{}\gamma_2}{\partial{}x_2}=\mathds{1}\cdot\gamma_i'\\
            \nonumber&x_3'=\frac{\partial{}\gamma_3}{\partial{}x_3}=\mathds{1}\cdot\gamma_i'\\
            \nonumber&x_4'=\frac{\partial{}\gamma_4}{\partial{}x_4}=\mathds{1}\cdot\gamma_i'
        \end{align}
        All the intermediate variables required for the computation at a certain node can be obtained from upstream.
        Line 1 of Eq.~\ref{eq:chain} (output node in Example~\ref{ex:backward}) is the conjugated Fourier coefficient computed in the preprocessing stage and hence, no floating point operations are needed.
        Lines 2-7 (intermediate nodes) are point-wise multiplications of two vectors.
        Lines 8-11 (input nodes) are taking the sum of column vectors.

        We can observe from Eq.~\ref{eq:chain} that, given a constraint with $k$ literal, there will be $2k-2$ intermediate nodes and $k$ input nodes.
        The computation at an intermediate node consists of $k+1$ multiplications.
        The computation at an input node consists of $k$ additions.
        The floating point operations of the overall procedure consist of $k^2$ additions and $2k^2-2$ multiplications.
        Therefore, the complexity of this algorithm scales at $O(k^2)$.

    \subsection{Proof of Proposition~\ref{prop:parallel}}
        We consider the graph traversal from Example~\ref{ex:forward} and Example~\ref{ex:backward} to show the proof but we generalize to any constraint $c$ with $k$ literals.
        The ideal execution time $O^*(\cdot)$ depends on the parallelizable component in the graph, which has a layer-wise topology.
        
        First, we analyze the graph traversal in Example~\ref{ex:forward}, which is related to the evaluation as in Appendix~\ref{app:forward}.
        At the input layer ($x_i$'s), the outer addition can run in $O^*(1)$ time.
        At the intermediate layers ($\gamma_i$'s), the computation within the node is the point-wise multiplication of two vectors, which can be obtained concurrently.
        However, the overall time depends on the graph topology, \emph{i.e.}, the depth of the binary tree is $\log{}k$.
        Thus, the best theoretical execution time to traverse the intermediate layers is $O^*(\log{}k)$.
        At the output layer, the vector-vector multiplication ($\tilde{f}_c\cdot\gamma_c$) runs in $O^*(\log{k})$.

        Next, we analyze the graph traversal in Eg.~\ref{ex:backward}, which is related to the differentiation as in Appx.~\ref{app:backward}.
        At the output layer, no floating point operations are required.
        The computation at the intermediate layers is similar to the left half of the graph, as well as the best theoretical execution time.
        At the input layer, the parallel sum reduction ($\mathds{1}\cdot\gamma_i'$) runs in $O^*(\log{k})$;

        Therefore, the ideal execution time of differentiating the Walsh expansion of a symmetric Boolean constraint with $k$ literals is $O^*(\log{k})$.

    \subsection{Proof of Corollory~\ref{cor:reduction}}
        \begin{lemma}
            (Euler's formula) 
            For any real number $x$:
            \begin{equation*}
                \exp\left(ix\right)=\cos(x)+i\sin(x)
            \end{equation*}
        \end{lemma}
        Given $\omega^{ab}=\exp\left(\frac{i2\pi{}ab}{k+1}\right)$, it can be transformed into $\cos\left(\frac{2\pi{}ab}{k+1}\right)+i\sin\left(\frac{2\pi{}ab}{k+1}\right)$.
        \begin{lemma}
            (Lagrange's trigonometric identities)
            Given $\theta\not\equiv{}0 (\mod{}2\pi)$: 
            \begin{align*}
                &\sum_{k=0}^n\sin{k\theta}=\frac{\cos\frac{\theta}{2}-\cos\left(\left(n+\frac{1}{2}\right)\theta\right)}{2\sin\frac{\theta}{2}}\\
                &\sum_{k=0}^n\cos{k\theta}=\frac{\sin\frac{\theta}{2}+\sin\left(\left(n+\frac{1}{2}\right)\theta\right)}{2\sin\frac{\theta}{2}}
            \end{align*}
        \end{lemma}
        Therefore, the coefficients in Eq.~\ref{eq:xor} can be written as:
        \begin{align}
            \nonumber
            \sum_{a=0}^k\omega^{ab}=&\sum_{a=0}^k\cos\left(\frac{2\pi{}ab}{k+1}\right)+i\sum_{a=0}^k\sin\left(\frac{2\pi{}ab}{k+1}\right)\\
            =&\frac{\sin\frac{\pi{}b}{k+1}-\sin\left(\pi{}b+\frac{\pi{}b}{k+1}\right)}{2\sin\frac{\pi{}b}{k+1}}
            +i\frac{\cos\frac{\pi{}b}{k+1}+\cos\left(\pi{}b+\frac{\pi{}b}{k+1}\right)}{2\cos\frac{\pi{}b}{k+1}}
            \label{eq:euler}
        \end{align}
        When $b\in\mathbb{Z}^+$, Eq.~\ref{eq:euler} is 0, \emph{i.e.}, all the computation related to $b\in\mathbb{Z}^+$ is trivial computation.
        
        When $b=0$, $\sum_{a=0}^k\omega^{ab}=k+1$.
        As a result, Eq.~\ref{eq:xor} will be simplified as ${WE}_{XOR}=\prod_{b=0}^kx_b$, \emph{i.e.}, purely multiplication.
        Thus, the complexity will be reduced to $O(k)$.

\end{document}